\documentclass{article}

\PassOptionsToPackage{square, numbers, compress}{natbib}

\usepackage[preprint]{neurips_data_2024}

\usepackage[utf8]{inputenc} % allow utf-8 input
\usepackage[T1]{fontenc}    % use 8-bit T1 fonts
\usepackage{hyperref}       % hyperlinks
\usepackage{url}            % simple URL typesetting
\usepackage{booktabs}       % professional-quality tables
\usepackage{amsfonts}       % blackboard math symbols
\usepackage{nicefrac}       % compact symbols for 1/2, etc.
\usepackage{microtype}      % microtypography
\usepackage{xcolor}         % colors

\usepackage{amsmath}        % equations
\usepackage{siunitx}        % SI units
\usepackage{graphicx}       % figures
\usepackage{multirow}       % table
\usepackage{array}          % table
\usepackage{subcaption}     % subfigures
\newcommand{\MANDataset}{MAN TruckScenes}
\DeclareSIUnit{\nothing}{\relax}

\title{\MANDataset{}: A multimodal dataset for autonomous trucking in diverse conditions}

\author{%
  Felix Fent\textsuperscript{1} \\
  \And
  Fabian Kuttenreich\textsuperscript{2}\thanks{corresponding author} \\
  \And
  Florian Ruch\textsuperscript{2} \\
  \And
  Farija Rizwin\textsuperscript{2} \\
  \AND
  Stefan Juergens\textsuperscript{2} \\
  \And
  Lorenz Lechermann\textsuperscript{2} \\
  \And
  Christian Nissler\textsuperscript{2} \\
  \And
  Andrea Perl\textsuperscript{2} \\
  \AND
  Ulrich Voll\textsuperscript{2} \\
  \And
  Min Yan\textsuperscript{2} \\
  \And
  Markus Lienkamp\textsuperscript{1} \\
  \AND
  \textsuperscript{1}Technical University of Munich \\
  School of Engineering \& Design \\
  Institute of Automotive Technology \\
  \And
  \textsuperscript{2}MAN Truck \& Bus SE \\
  \texttt{truckscenes@man.eu} \\
}

\begin{document}

\maketitle

\begin{abstract}
  Autonomous trucking is a promising technology that can greatly impact modern logistics and the environment. Ensuring its safety on public roads is one of the main duties that requires an accurate perception of the environment. To achieve this, machine learning methods rely on large datasets, but to this day, no such datasets are available for autonomous trucks. In this work, we present \MANDataset{}, the first multimodal dataset for autonomous trucking. \MANDataset{} allows the research community to come into contact with truck-specific challenges, such as trailer occlusions, novel sensor perspectives, and terminal environments for the first time. It comprises more than 740 scenes of 20\,s each within a multitude of different environmental conditions. The sensor set includes 4 cameras, 6 lidar, 6 radar sensors, 2 IMUs, and a high-precision GNSS. The dataset's 3D bounding boxes were manually annotated and carefully reviewed to achieve a high quality standard. Bounding boxes are available for 27 object classes, 15 attributes, and a range of more than 230\,m. The scenes are tagged according to 34 distinct scene tags, and all objects are tracked throughout the scene to promote a wide range of applications. Additionally, \MANDataset{} is the first dataset to provide 4D radar data with 360° coverage and is thereby the largest radar dataset with annotated 3D bounding boxes. Finally, we provide extensive dataset analysis and baseline results. The dataset, development kit, and more are available online.
\end{abstract}

\section{Introduction}
\label{sec:into}
Autonomous trucking has the potential to fundamentally change today's traffic by increasing safety on public roads, reducing logistics costs, and counteracting the shortage of drivers~\cite{Bray2022, Schuster2023}. However, the safe and reliable operation of autonomous trucks depends on an accurate perception of the surroundings. To achieve this, modern self-driving vehicles rely on machine learning algorithms to detect, track, and predict surrounding objects. However, the use of machine learning methods also drives the need for large-scale datasets.

While numerous datasets exist for autonomous passenger cars~\cite{Liu2024}, datasets for autonomous trucks are missing. However, heavy-duty vehicles have their unique challenges. Large vehicles require different sensor mounting positions and rely on multiple sensors to cover the entire surrounding area. Moreover, trucks have to contend with occlusions from their own vehicle that change dynamically due to a movable truck-trailer combination and are affected by relative movements between their chassis and cabin. Furthermore, long-haul trucks operate in inherently different surroundings, such as logistics or container terminals and their functionality has to be ensured under all environmental conditions to maintain a functioning logistics system. Therefore, a dedicated truck dataset is needed to develop reliable perception solutions for self-driving trucks under all conditions.

\begin{figure}[tb]
    \centering
        % Lidar
        \begin{subfigure}{0.326\linewidth}
            \includegraphics[width=4.55cm]{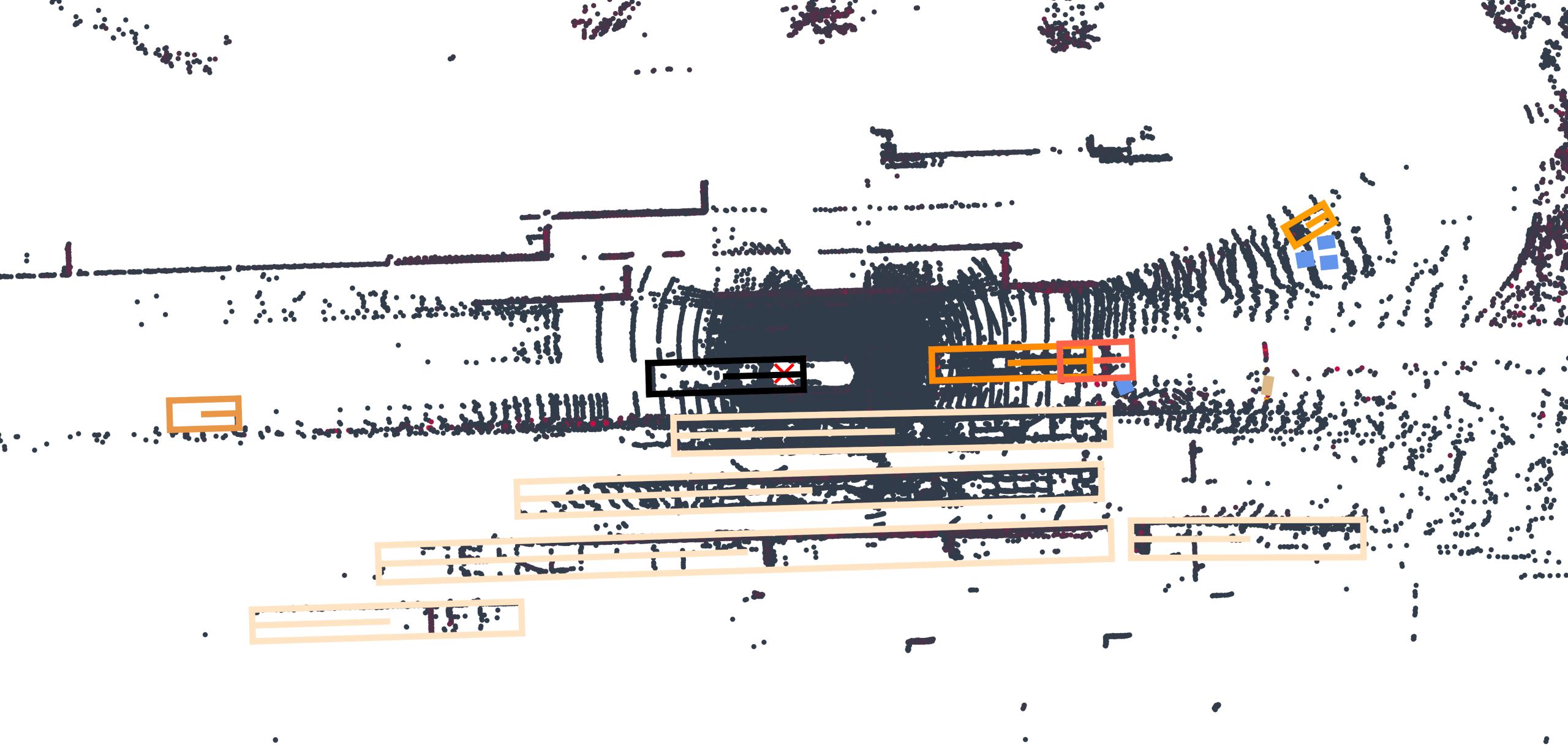}
        \end{subfigure}
        \hfil
        \begin{subfigure}{0.326\linewidth}
            \includegraphics[width=4.55cm]{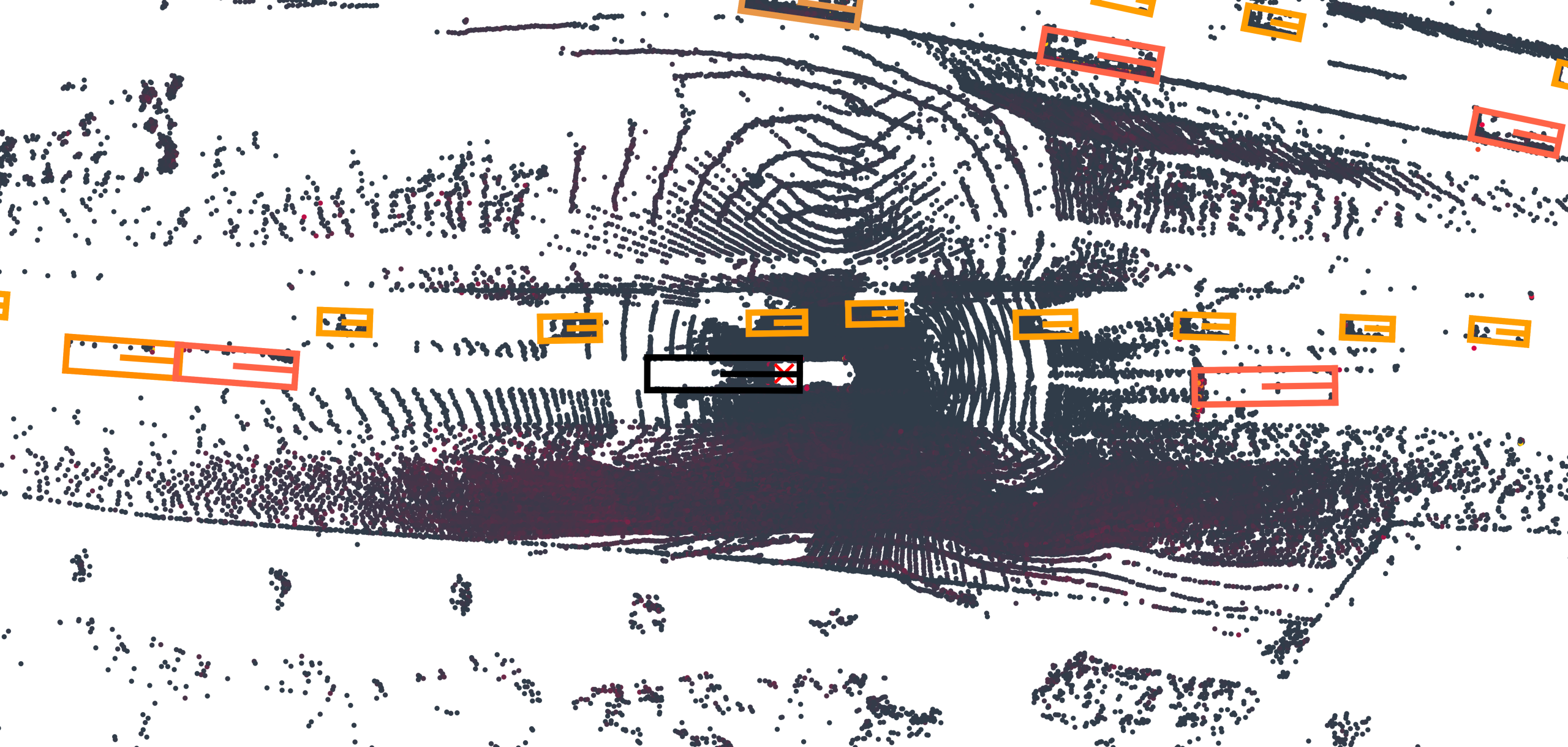}
        \end{subfigure}
        \hfil
        \begin{subfigure}{0.326\linewidth}
            \includegraphics[width=4.55cm]{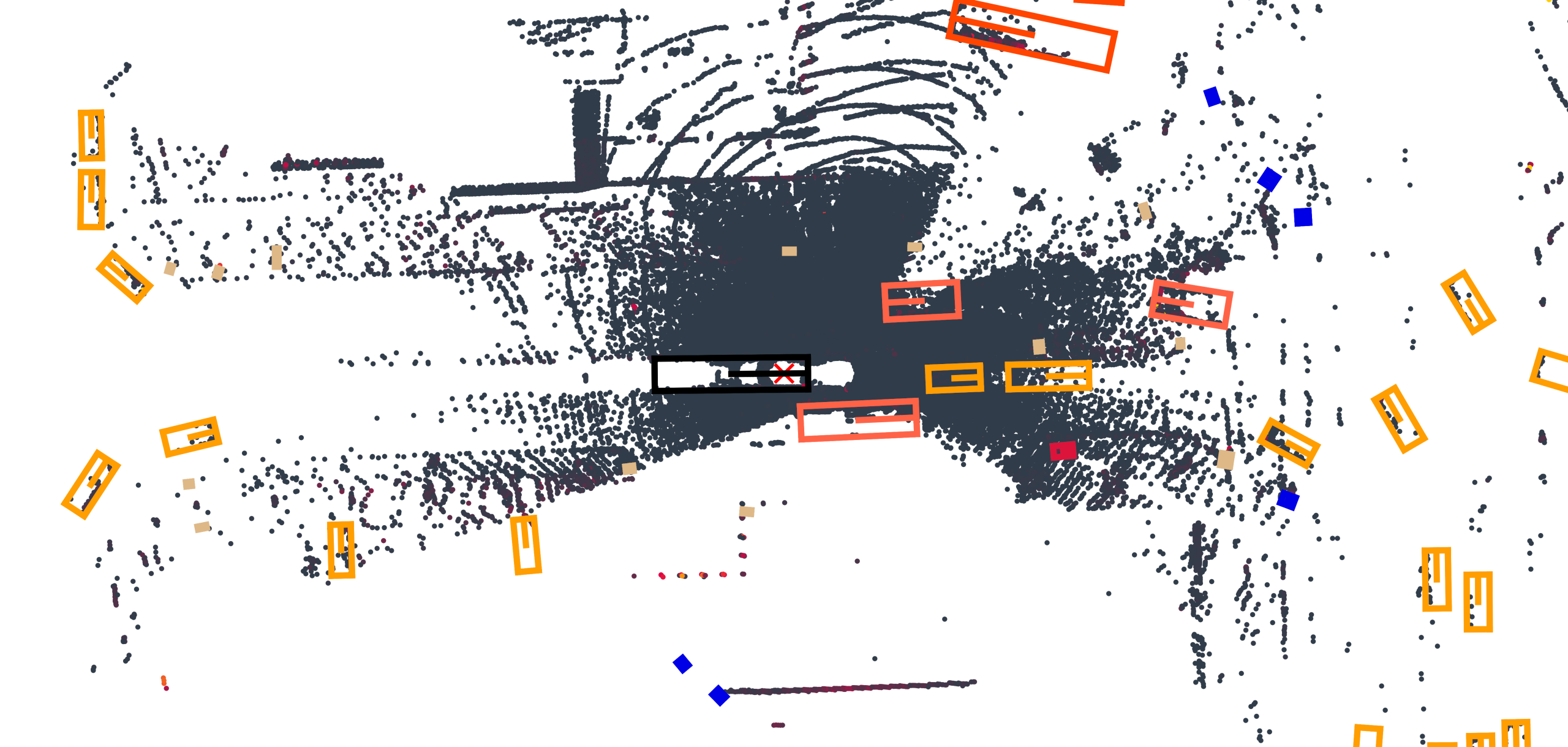}
        \end{subfigure}
        \hfil

        % Camera
        \begin{subfigure}{0.326\linewidth}
            \includegraphics[width=4.55cm]{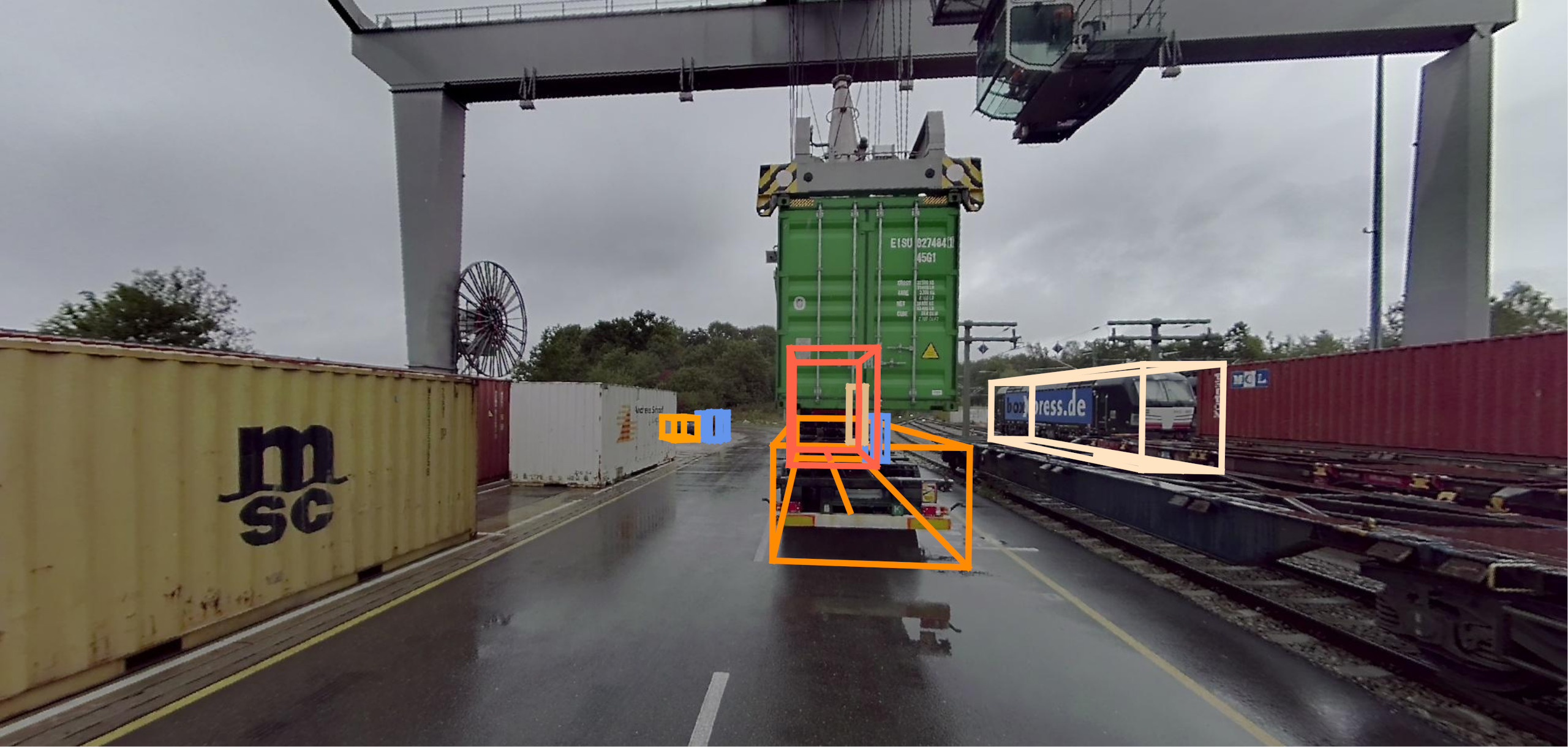}
        \end{subfigure}
        \hfil
        \begin{subfigure}{0.326\linewidth}
            \includegraphics[width=4.55cm]{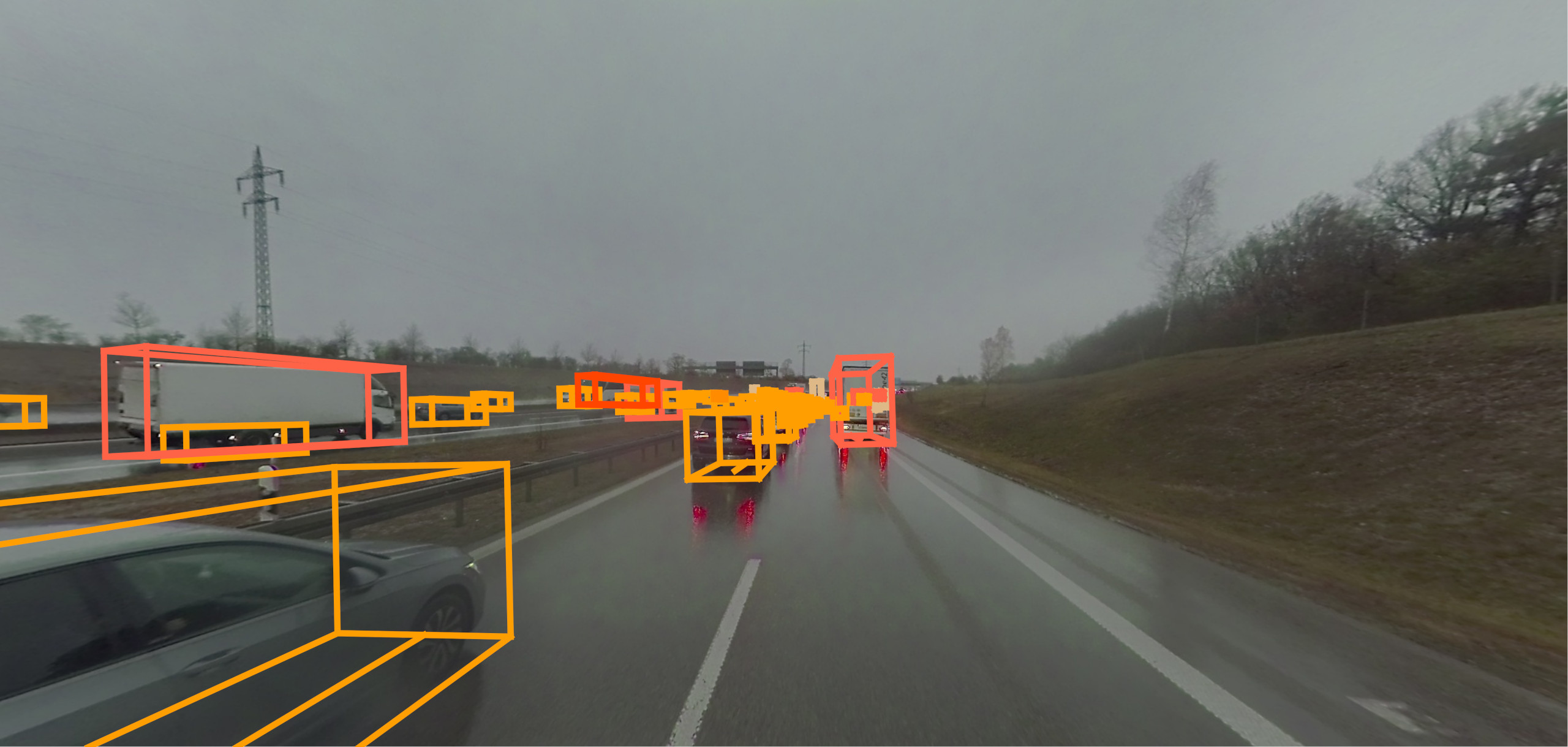}
        \end{subfigure}
        \hfil
        \begin{subfigure}{0.326\linewidth}
            \includegraphics[width=4.55cm]{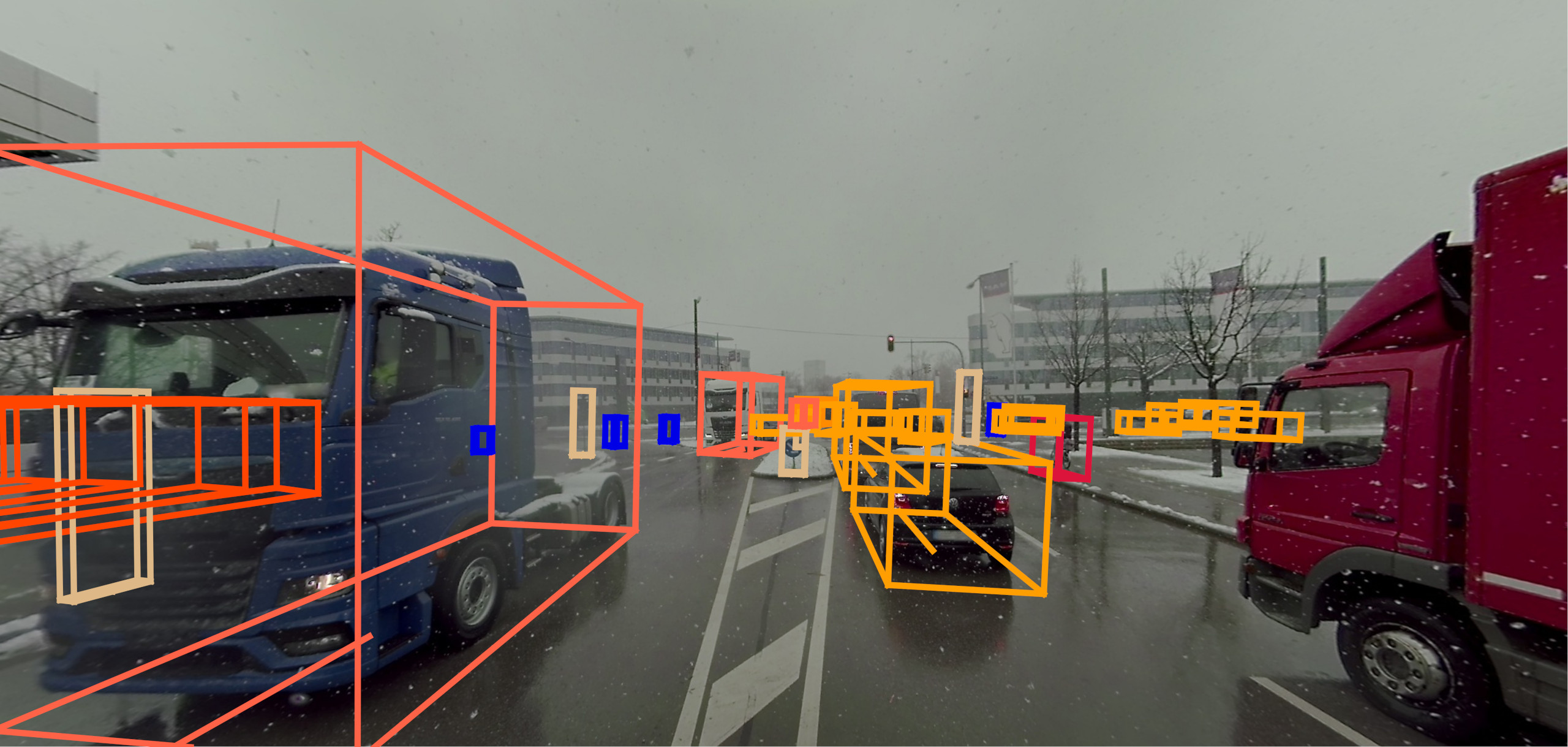}
        \end{subfigure}
        \hfil

        % Radar
        \begin{subfigure}{0.326\linewidth}
            \includegraphics[width=4.55cm]{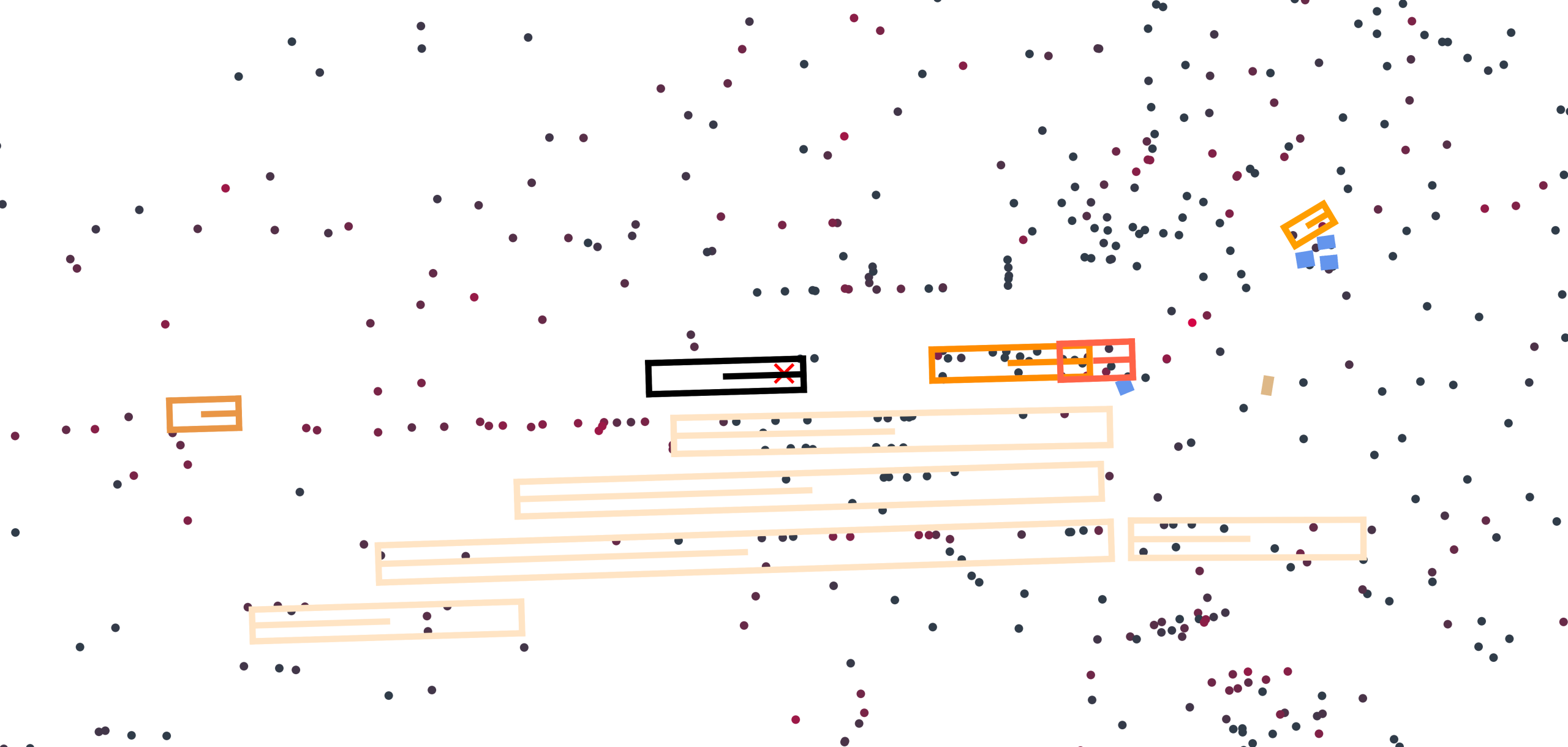}
            \caption{terminal}
        \end{subfigure}
        \hfil
        \begin{subfigure}{0.326\linewidth}
            \includegraphics[width=4.55cm]{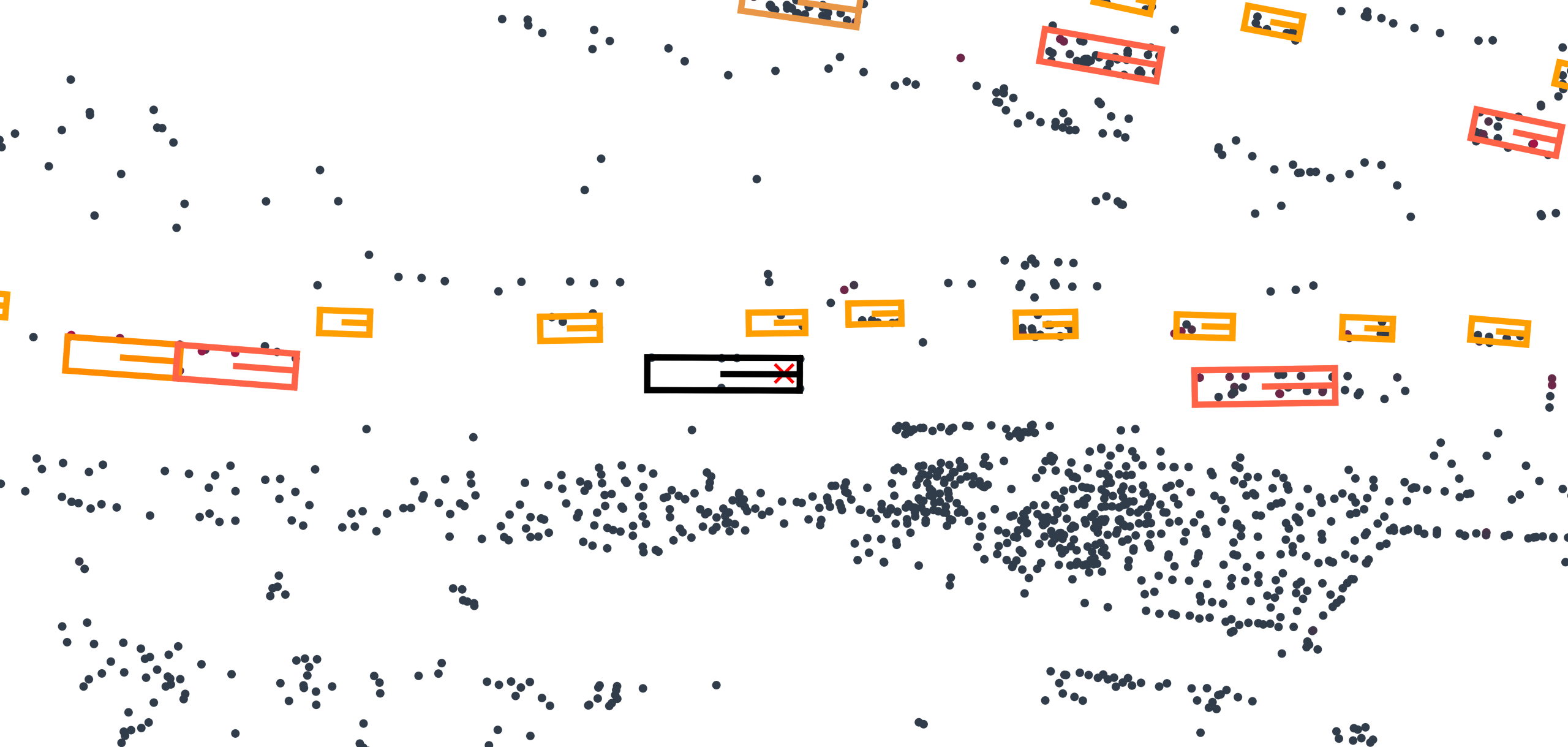}
            \caption{rain}
        \end{subfigure}
        \hfil
        \begin{subfigure}{0.326\linewidth}
            \includegraphics[width=4.55cm]{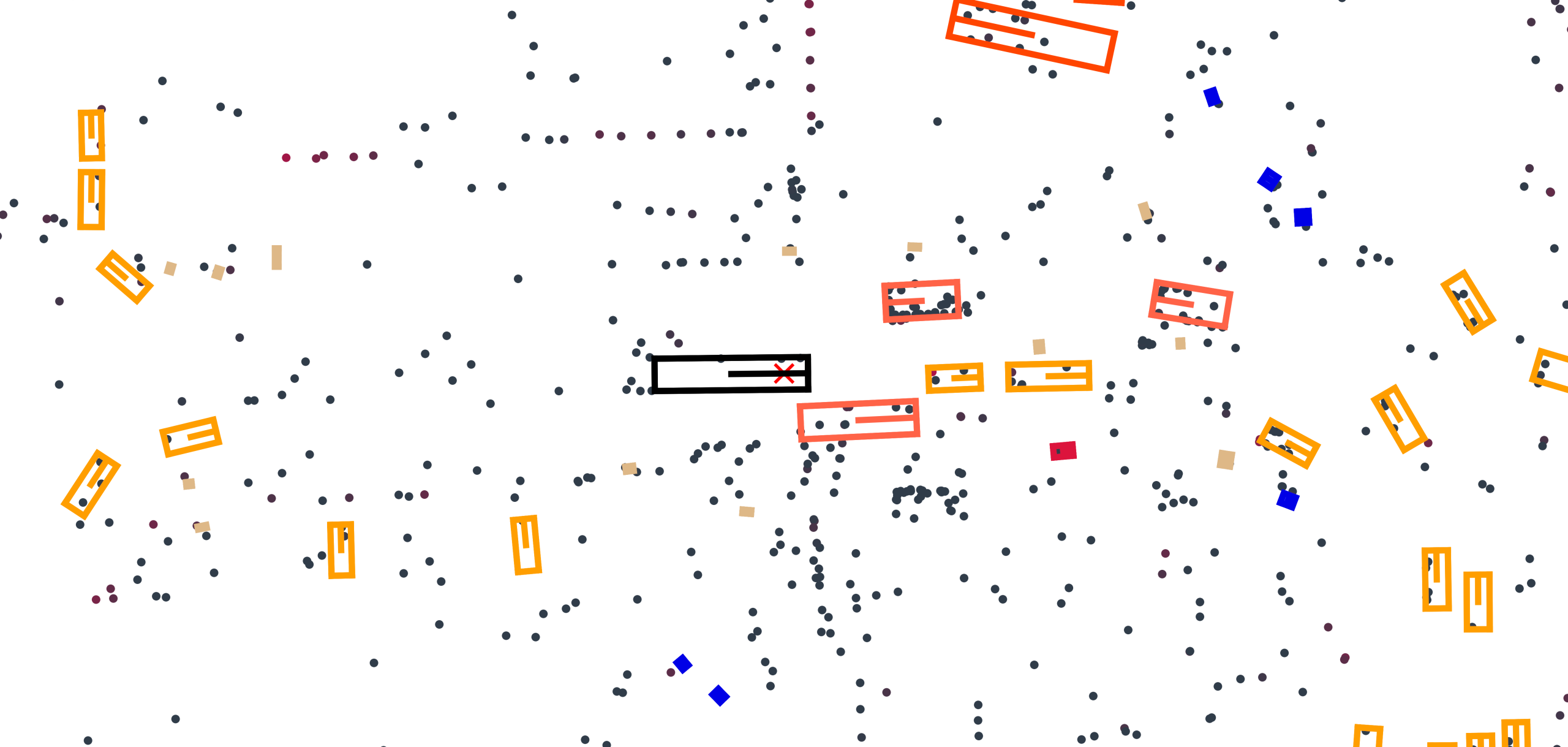}
            \caption{snow}
        \end{subfigure}
        \hfil
      
    \caption{Exemplary selection of a terminal, rain, and snow scene of the \MANDataset{} dataset. The top row shows the fused lidar point cloud, the center row shows images of the front left camera, and the fused radar point cloud is shown at the bottom.}
    \label{fig:opening}
\end{figure}

To address this research gap and accelerate the development of self-driving trucks, we present \MANDataset{}, the first large-scale dataset for autonomous trucking. Our multimodal dataset comprises data from a state-of-the-art sensor suite, including multiple high-resolution cameras, lidar, and radar sensors to provide full coverage of the surrounding area. Especially, the inclusion of six 4D radar sensors makes it the largest radar dataset available. In addition, the data of a high-precision GNSS and two IMU units are included to support a multitude of applications.

Moreover, most existing AV datasets are restricted to a particular operational design domain, whereas the \MANDataset{} dataset covers a large geographical area, three seasons, nighttime driving, and numerous different weather conditions, including fog, rain, and snow. Furthermore, the dataset includes scenes in logistics terminals and recordings of vehicles with high relative velocities on the German Autobahn, making it the first dataset to promote the unique challenges of long-haul trucks.

In order to support the development of deep learning methods, \MANDataset{} provides high-quality annotations for 747 scenes. These annotations consist of manually annotated 3D bounding boxes with unique instance identifiers for object tracking, standardized scene tags, and attribute labels to indicate the object's state. The dataset annotation was subject to a multi-stage labeling and quality assurance process of specially trained labelers to ensure accurate and consistent labeling.

The dataset will be published alongside a development kit, evaluation code, taxonomy, and detailed annotation instructions. This ensures transparency and reproducibility and simplifies the use of the dataset. Furthermore, we build on the established nuScenes~\cite{Caesar2020} data format to ensure easy integration and code compatibility. Finally, \MANDataset{} is published under the CC~BY-NC-SA~4.0 license to accelerate research on autonomous trucks and shape the future of logistics.

% Motivation for autonomous trucking
% \begin{itemize}
%     \item Driver shortage
%     \item Driving time
%     \item Driver cost
%     \item Efficiency
%     \item Safety
% \end{itemize}

% Datasets exist, but special challenges of autonomous trucks
% \begin{itemize}
%     \item Different sensor positions
%     \item Limited field of view (vehicle dimensions, occlusion)
%     \item Movable cabin
%     \item Special ODDs
% \end{itemize}

% Contributions
% \begin{itemize}
%     \item First dataset for autonomous trucking.
%     \item Multimodal dataset
%     \item Environmental conditions
%     \item Special ODDs
%     \item Benchmark, devkit, specifications
%     \item High-quality annotations
%     \item High relative velocities (German highway)
% \end{itemize}

\section{Related Work}
\label{sec:sota}
The advances in autonomous driving have largely been driven by the release of public datasets. Over the course of the last decade, we have seen a trend towards high-quality, large-scale, and more diverse multimodal datasets that enabled innovation in the autonomous driving domain.

\begin{table}
  \caption{Comparison of publicly available perception datasets for autonomous driving. The coverage represents the geographical coverage calculated according to \cite{Sun2020} and range refers to the 99.9th percentile of all bounding box distances~\cite{Alibeigi2023}. Scenes are temporal consistent sequences and samples are annotated keyframes, following the nuScenes~\cite{Caesar2020} notation. Vehicle refers to the recording vehicle.}
  \label{tab:comparison}
  \centering
  \begin{tabular}{lcccccccc}
    \toprule
    & \begin{tabular}[c]{@{}c@{}}KITTI\\ \cite{Geiger2013}\end{tabular} & \begin{tabular}[c]{@{}c@{}}nuScenes\\ \cite{Caesar2020}\end{tabular} & \begin{tabular}[c]{@{}c@{}}Waymo\\ \cite{Sun2020}\end{tabular} & \begin{tabular}[c]{@{}c@{}}Argo2\\ \cite{Wilson2021}\end{tabular} & \begin{tabular}[c]{@{}c@{}}ZOD\\ \cite{Alibeigi2023}\end{tabular} & \begin{tabular}[c]{@{}c@{}}aiMotive\\ \cite{Matuszka2023}\end{tabular} & \begin{tabular}[c]{@{}c@{}}VoD\\ \cite{Palffy2022}\end{tabular} & \begin{tabular}[c]{@{}c@{}}Ours\\ ~\end{tabular} \\
    \midrule
    Scenes & 22 & 1000 & 1150 & 1000 & 1473 & 176 & 21 & 747 \\
    Sample & \SI{1.5}{\kilo\nothing} & \SI{40}{\kilo\nothing} & \SI{230}{\kilo\nothing} & \SI{150}{\kilo\nothing} & \SI{1.5}{\kilo\nothing} & \SI{27}{\kilo\nothing} & \SI{9}{\kilo\nothing} & \SI{30}{\kilo\nothing} \\
    \midrule
    Duration & \SI{1.5}{\hour} & \SI{5.5}{\hour} & \SI{6.4}{\hour} & \SI{4.2}{\hour} & \SI{8.2}{\hour} & \SI{0.7}{\hour} & \SI{0.2}{\hour} & \SI{4.2}{\hour} \\
    Coverage & - & \SI{4}{\kilo\metre\squared} & \SI{76}{\kilo\metre\squared} & \SI{17}{\kilo\metre\squared} & \SI{26}{\kilo\metre\squared} & \SI{180}{\kilo\metre\squared} & \SI{2}{\kilo\metre\squared} & \SI{100}{\kilo\metre\squared} \\
    \midrule
    Camera & 4 & 6 & 5 & 9 & 1 & 4 & 1 & 4 \\
    Lidar & 1 & 1 & 5 & 2 & 3 & 1 & 1 & 6 \\
    Radar & 0 & 5 & 0 & 0 & 0 & 2 & 1 & 6 \\
    \midrule
    GNSS & 1 & 1 & 0 & 0 & 2 & 1 & 1 & 1 \\
    IMU & 1 & 1 & 0 & 0 & 2 & 1 & 0 & 2 \\
    Map & no & yes & no & yes & no & no & no & no \\
    \midrule
    Range & \SI{91}{\metre} & \SI{141}{\metre} & \SI{80}{\metre} & \SI{214}{\metre} & \SI{245}{\metre} & \SI{228}{\metre} & \SI{26}{\metre} & \SI{226}{\metre} \\
    Classes & 3 & 23 & 4 & 30 & 15 & 14 & 13 & 27 \\
    \midrule
    Vehicle & car & car & car & car & car & car & car & truck \\
    \bottomrule
  \end{tabular}
\end{table}

The KITTI~\cite{Geiger2013} dataset, released in 2012, is one of the most influential autonomous driving datasets to date. It provides 22 scenes of annotated camera and lidar data in combination with high-precision GNSS and IMU data. However, the dataset's extent and diversity are limited, radar and map data are not provided, and driving data under severe weather conditions are not included.

The nuScenes~\cite{Caesar2020} dataset had a great impact as one of the first large-scale, multimodal datasets for autonomous driving. Next to camera and lidar data, it also includes radar and map data for 1000 annotated scenes. It provides data from two different cities, annotates 23 different object classes, and introduces unlabeled intermediate sensor data sweeps. In addition to the small geographical coverage and limited coverage of severe weather conditions, the nuScenes dataset has been criticized for its sparse radar data~\cite{Engels2021, Schumann2021}.

The Waymo Open~\cite{Sun2020} dataset is one of the largest annotated datasets for autonomous driving with a focus on scalability. It provides annotated camera and lidar data for 1150 scenes, covers a geographical area of \SI{76}{\kilo\metre\squared}, and provides a rich ecosystem of different related datasets. However, it does not include radar, GNSS, or map data and has a limited annotation range of \SI{80}{\metre} with only four annotation classes.

The Argoverse~2~\cite{Wilson2021} dataset takes a different approach with the provision of long-range annotations, comprehensive map data, and extensive object taxonomy. Nevertheless, it does not provide radar, GNSS, or IMU data and covers a limited geographical area.

The Zenseact Open Dataset~(ZOD)~\cite{Alibeigi2023} makes a different set of trade-offs. It provides 1473 annotated scenes from six different countries and facilitates long-range perception with an annotation range of \SI{245}{\metre}. Despite its provision of two GNSS and IMU units, the ZOD's sensor setup, as well as its number of annotated samples, is limited.

The aiMotive~\cite{Matuszka2023} dataset focuses on the collection of diverse driving scenes under severe weather conditions. It provides long-range annotations and data from a multimodal sensor setup for three distinct areas. However, the dataset's extent is limited to a duration of \SI{0.7}{\hour} and the two radar sensors do not cover a \SI{360}{\degree} field of view (FoV).

The View-of-Delft~\cite{Palffy2022} dataset is one of the first large-scale datasets with annotated 4D radar data and promotes the development of radar-based perception methods for vulnerable road users (VRUs) in city environments. Nevertheless, the geographical coverage of the dataset is limited, it does not include severe weather conditions, and only provides data from a single front-facing radar sensor.

While there are numerous other datasets focusing on various different aspects~\cite{Liu2024}, all of the aforementioned datasets are limited to passenger cars, as shown in Table~\ref{tab:comparison}. To the best of our knowledge, there are only two datasets that include truck data. One is the TuSimple~\cite{Yoo2020} dataset, which consists of 6408 images from a single front camera and provides annotations for lane markings only. The other is the SurMine~\cite{Song2023} dataset, which is a proprietary dataset that only includes 2D bounding box annotations for 10665 camera images recorded in a mining facility. Hence, there are no large-scale or multimodal perception datasets for autonomous trucking.

% Structure
% \begin{enumerate}
%   \item Comparison table
%   \item Autonomous driving datasets
%   \item Autonomous truck datasets
% \end{enumerate}

\section{Dataset}
\label{sec:dataset}
\MANDataset{} aims to close this research gap by providing the first large-scale multimodal dataset for autonomous trucking. It consists of 747 scenes from typical long-haul truck environments and provides multimodal data with long-range annotations based on the nuScenes~\cite{Caesar2020} format.

\subsection{Sensor Setup}
\label{sec:setup}
\begin{figure}
  \centering
  \includegraphics[width=13.97cm]{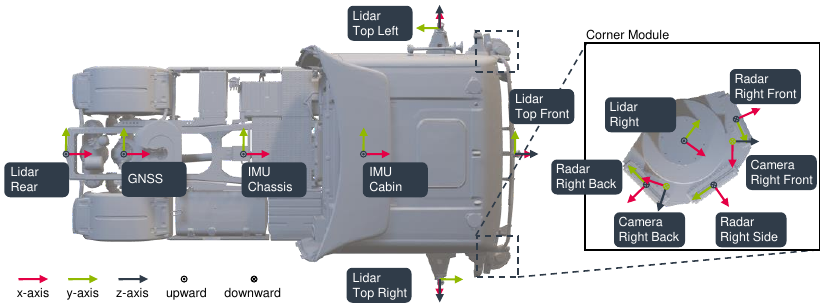}
  \caption{Placement of the sensors and their corresponding coordinate systems from a top view perspective. The right sensor module is shown in the detailed drawing, the left module is identical but mirrored and the radar sensors are flipped around the x-axis. The top-mounted lidar sensors (at the sides and the front) are tilted downward. The GNSS uses a virtual coordinate system similar to the vehicle frame but is located next to the chassis IMU.}
  \label{fig:sensors}
\end{figure}

The sensor suite consists of a multimodal sensor setup with four cameras, six lidar, and six radar sensors. Additionally, the dataset provides high-precision RTK-GNSS data and measurements of two IMU units. Detailed information on the sensor specifications can be found in Table~\ref{tab:sensors} and the sensor positions are shown in Figure~\ref{fig:sensors}.

\begin{table}
  \caption{Sensor specifications of the \MANDataset{} setup.}
  \label{tab:sensors}
  \centering
  \begin{tabular}{ll}
    \toprule
     Sensor & Details \\
    \midrule
    Camera & 4\texttimes{} Sekonix SF3324, RGB, \SI{10}{\hertz}, 1928 \texttimes{} 1208, \SI{120}{\degree} \texttimes{} \SI{73}{\degree} FoV \\
    Lidar & 2\texttimes{} Hesai Pandar64, \SI{10}{\hertz}, 64 layer, \SI{360}{\degree} \texttimes{} \SI{40}{\degree} FoV, \SI{200}{\metre}@\SI{10}{\percent} \\
          & 4\texttimes{} Ouster OS0, \SI{10}{\hertz}, 64 layer, \SI{360}{\degree} \texttimes{} \SI{90}{\degree} FoV, \SI{35}{\metre}@\SI{10}{\percent} \\
    Radar & 6\texttimes{} Continental ARS 548 RDI, \SI{20}{\hertz}, \SI{76}{\giga\hertz}, \SI{100}{\degree} \texttimes{} \SI{28}{\degree} \\
    GNSS & 1\texttimes{} GeneSys ADMA-G-PRO+, \SI{100}{\hertz}, \SI{0.01}{\metre} pos.,  \SI{0.015}{\degree} heading \\
    IMU & 2\texttimes{} Xsens MTi-680G-SK, \SI{100}{\hertz}, 9 DoF \\
    \bottomrule
  \end{tabular}
\end{table}

The main perception sensors are arranged in two sensor modules, one on either side of the vehicle, to maximize spatial coverage and minimize relative sensor movements. Each sensor module houses two Sekonix cameras, one Hesai lidar, and three Continental radar sensors. Besides these two "corner modules", the vehicle is equipped with three Ouster lidar sensors mounted on the roof of the cabin, which are tilted downwards for blindspot coverage, and one additional Ouster lidar at the rear of the semi-trailer truck. The top mounted lidar sensors and the lidar sensors of the corner modules are positioned at a height of \SI{3.2}{\metre} and \SI{2.2}{\metre}, respectively. These elevated positions help to reduce occlusions, protect pedestrians, and prevent sensor damage. However, this sensor perspective marks a significant difference from conventional passenger car datasets.

Another unique feature of our dataset is the inclusion of six 4D radar sensors with a spatial coverage of nearly \SI{360}{\degree} (except from the occlusion by the ego vehicle's trailer). In contrast to conventional 3D radar sensors that operate in the range-azimuth plane, 4D radar sensors can resolve objects in both azimuth and elevation angle. Furthermore, our radar sensors provide on average 2600 points per sample, while conventional radar sensors (e.g.\ in the nuScenes dataset) provide only 200 data points. Therefore, \MANDataset{} is the first dataset to provide 4D radar data with \SI{360}{\degree} coverage and is the largest radar dataset with annotated 3D bounding boxes.

To accurately capture the vehicle's state and position, the data of two IMUs and one RTK-GNSS are included. The GNSS unit uses a dual antenna setup to measure the heading angle of the vehicle and RTK correction data to achieve a position accuracy of up to \SI{0.01}{\metre}. The two IMU units are mounted on the chassis and the cabin of the vehicle to get accurate measurements of the overall vehicle state on one side and measure relative movements between the chassis and the cabin on the other. This is important to compensate for sensor movements that are primarily mounted on the movable cabin and represent a special challenge of the truck case.

\subsection{Sensor Synchronization}
\label{sec:sync}
Sensor synchronization is an important topic for multimodal datasets to ensure temporal sensor data alignment. This includes not only sensor time synchronization, which ensures that all sensors are based on the same reference clock, but also triggering the sensors to achieve cross-modality consistency.

The sensor time synchronization is based on the Precision Time Protocol (PTP) in accordance with the IEEE/IEC 1588-2008~\cite{IEEE1588}. This procedure ensures the alignment of the individual sensor clocks with a high-precision Mobatime DTS 4160 PTP grandmaster clock. As a result, the individual sensors have a time deviation of less than \SI{100}{\micro\second} and are referenced to the global UTC time. However, this only ensures that all sensors are based on the same reference clock, but not that all measurements are taken at the exact same point in time.

To temporally align the actual sensor measurements, the sensors have to be triggered. For this purpose, the lidar sensors are defined as reference sensors and all other perception sensors are triggered based on them. In the first step, all cabin-mounted lidar sensors are phase-synchronized such that their lidar points are temporally consistent in a clockwise manner. Secondly, the four cameras are triggered at the point where the lidar sweep and the rolling shutter of the cameras align in the center of the image. Lastly, the radar sensors are synchronized with the corresponding lidar sensors but triggered with a small delay to one another to minimize interference between them. Besides that, all radar sensors are assigned to slightly different frequency bands to further minimize interference.

% Provision of Lidar timestamps

\subsection{Sensor Calibration}
\label{sec:calib}
While sensor synchronization ensures the temporal alignment of the sensor data, sensor calibration ensures the spatial alignment. The calibration aims to determine both the extrinsic and intrinsic sensor parameters to estimate both the sensor poses as well as the parameters of the sensor models.

The sensor calibration consists of four steps and includes an initial extrinsic calibration, a combined extrinsic and intrinsic calibration, an angular correction, and a final validation. The first step is a photogrammetric scan of the vehicle to determine the position and orientation of the sensors with respect to the vehicle frame. The vehicle frame is chosen to be the center of the rear axle projected onto the ground plan in accordance with ISO~8855:2011~\cite{ISO8855}. For the main calibration, the vehicle is placed in a dedicated calibration hall equipped with specialized calibration targets. These targets are represented by differently oriented planes in 3D space equipped with unique identifiers. The actual calibration uses a plane-matching algorithm to jointly estimate the extrinsic parameters of the camera and lidar sensors as well as the intrinsic camera parameters. Within this calibration procedure, the vehicle is moved back and forth to calibrate the sensors with respect to the vehicle frame. The third step uses a faraway calibration target aligned with the vehicle's longitudinal axis to correct remaining yaw angle errors and a dynamic landmark-based calibration to align the heading angle of the overall sensor setup. Finally, the calibration is validated on the data level by comparing point cloud alignments and projecting the point cloud data onto the image data. Moreover, an application-level validation is used to compare detections, odometry, and localization measurements to validate the calibration.

This procedure results in a precise calibration with pixel-level accuracy between camera and lidar sensors. However, the radar calibration solely relies on external measurements of the photogrammetric scan with a measurement tolerance of \SI{0.05}{\milli\metre}. Therefore, angular errors of up to \SI{0.03}{\degree} and sensor internal misalignments cannot be excluded. Besides that, the odometry-based validation can only ensure a heading error of less than \SI{0.1}{\degree} and one has to be aware of increasing uncertainties within the camera intrinsic calibration towards the edges of the camera image.

\subsection{Sensor Data}
\label{sec:data}
The provided sensor data was going through different processing steps to ensure compatibility with the nuScenes~\cite{Caesar2020} format and enhance usability. The properties of the five different sensor data types are explained in the following.

The camera data of the four camera sensors undergo an undistortion process using a pinhole camera model and a Lanczos interpolation scheme with a $8\times8$ kernel and constant padding values. The resulting image is cropped to a $1980\times943$ pixel size and stored as a compressed JPEG~\cite{ISO15444} image. The Camera data is sampled at a frequency of \SI{10}{\hertz} in correspondence with the lidar.

The lidar data is provided as point clouds of a variable number of points represented by their $x, y, \text{and}\, z$-coordinates in a cartesian sensor coordinate system, an intensity value, and an individual UNIX timestamp, allowing for point-wise motion compensation. The final point clouds are stored as binary (Marc Lehmann’s LZF) compressed data in the point cloud data (pcd) format to save on disc space.

The radar data is also represented as point clouds where each point is defined by its $x, y, \text{and}\, z$-coordinates, its relative radial velocity in $x, y, \text{and}\, z$-direction as well as its radar cross section (rcs). In contrast to the lidar data, the radar data is stored as pcd files in a binary format due to its smaller file size.

The IMU data includes the current velocity and acceleration in $x, y, \text{and}\, z$-direction as well as the angle and angular velocity (rate) in roll, pitch, and yaw. The data is stored in a JSON file. The GNSS data provides the vehicle's position in UTM-WGS84 coordinates mapped to cell U32 and orientation given as quaternion ($qw, qx, qy, qz$). The GNSS data off all timestamps is stored in a JSON file. IMU and GNSS data are sampled with a frequency of \SI{100}{\hertz} to provide the possibility for highly accurate motion estimation.

Besides that, the rear lidar is limited to an FoV of \SI{180}{\degree}, which increases its detection range to \SI{42}{\metre} at \SI{10}{\percent} reflectivity, and the points of the top mounted lidar sensors are cut off for yaw angles beyond $\pm$\SI{120}{\degree}. It is also worth mentioning that the sensor data is not just provided at the sample annotation frequency of \SI{2}{\hertz} but at their individual sensor captioning frequencies listed in Table~\ref{tab:sensors}. These unannotated intermediate frames are denoted as sweeps. It is also important to note that the radar sensors cannot guarantee a fixed sampling rate of \SI{20}{\hertz} but rather drop frames, if their internal data processing takes too long, which results in an average data rate of \SI{19.6}{\hertz}.

% Aspects
% \begin{itemize}
%     \item Undistorted camera data
%     \item JPG camera data
%     \item Radar data (x, y, z, etc.)
%     \item Lidar data (x, y, z, i, time)
%     \item Lidar rear 180° (hardware), lidar top 240° (software)
%     \item GNSS data (ego pose)
%     \item IMU data
%     \item Unlabeled data sweeps
% \end{itemize}

\subsection{Scene Selection}
\label{sec:selection}
\begin{figure}
  \centering
  \includegraphics[width=13.97cm]{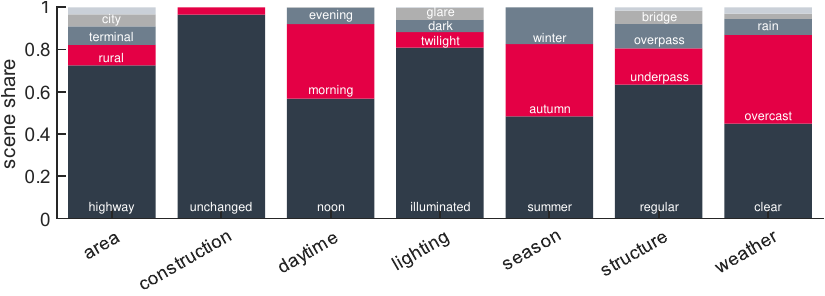}
  % \fbox{\rule[-.5cm]{0cm}{4cm} \rule[-.5cm]{12cm}{0cm}}
  \caption{Distribution of the scene tags for all 747 dataset scenes.}
  \label{fig:tags}
\end{figure}

The scene selection aims to select a diverse set of scenarios with challenging driving situations representing long-haul trucks' operational design domain. To meet this goal, 747 scenes with approximately \SI{20}{\second} each are manually selected from more than \SI{25}{\hour} of measurement drives. The scenes include recordings from various different areas (e.g.\ highway, terminal, rural, city), three seasons of the year, challenging weather situations (e.g.\ rain, fog, snow), and difficult environmental conditions (e.g.\ nighttime, twilight). Furthermore, the dataset covers different driving maneuvers (e.g.\ overtaking, offloading), traffic scenarios, and observations of rare object classes (e.g.\ animals, emergency vehicles). As a result, the dataset includes rainy nighttime drives that cause mirroring effects in the lidar point cloud, recordings in a logistics terminal with a lot of metal shipping containers which lead to multipath reflections in the radar point cloud, and tunnel passages with strong illumination changes that effect the camera images. Therefore, the dataset provides challenging conditions for all sensor modalities to promote research in the area of robust perception. The distribution of the different scene tags is shown in Figure~\ref{fig:tags}

% Aspects
% \begin{itemize}
%     \item Manually selected from a large database
%     \item Diverse scenarios (area, weather, season, daytime)
%     \item Challenging scenarios (construction sides, )
%     \item Truck specific (distribution highway, terminal, rural)
%     \item Coverage of all classes
%     \item Spatial coverage
% \end{itemize}

\subsection{Data Annotation}
\label{sec:annotation}
The data of the selected scenes is manually annotated at \SI{2}{\hertz} based on a fused and ego motion-compensated lidar point cloud aggregation. To ensure a high-quality standard, independent annotation and quality assurance companies have been commissioned. Within the labeling process, each sample undergoes a maximum of three consecutive annotation and quality assurance cycles until the quality target is met. Furthermore, randomly selected samples pass through a dual-control cycle to validate the annotation quality.

\begin{figure}
  \centering
  \includegraphics[width=13.97cm]{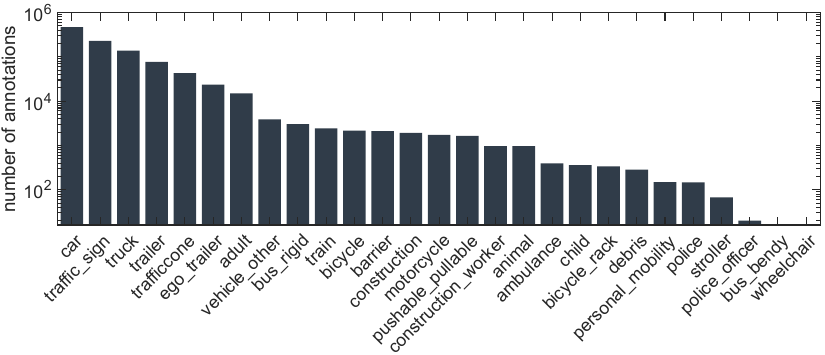}
  \caption{Number of annotated 3D bounding boxes for all 27 object classes across all dataset splits.}
  \label{fig:classes}
\end{figure}

Annotations are made in the form of 3D bounding boxes defined by their center point $(x, y, z)$, size $(w, l, h)$, and orientation $(qw, qx, qy, qz)$, given as quaternion. Each bounding box instance is classified according to 27 different object classes, using a hierarchical class structure, and in accordance with the nuScenes~\cite{Caesar2020} data format. The distribution of the object categories is shown in Figure~\ref{fig:classes}. In addition, every individual bounding box is labeled according to five attributes (with a total of 15 possible values) representing its visibility level or activity state. Furthermore, objects are tracked throughout the scene and assigned a unique and consistent tracking ID. Besides that, we label all scenes according to 34 distinct scene tags divided into seven different categories, including area, weather, and lighting conditions, as shown in Figure~\ref{fig:tags}.

% Aspects
% \begin{itemize}
%     \item Sample annotation at 2Hz
%     \item Labeling of 27 different object classes
%     \item Hierarchical label structure
%     \item Labeling of 15 different attributes in 5 categories
%     \item Labeling of 34 different scene tags in 7 categories
%     \item Labeling of tracking IDs
%     \item Labeling of 3D bounding boxes (position, rotation, size)
%     \item Quality assurance
% \end{itemize}

\subsection{Dataset Splits}
\label{sec:splits}
The dataset is split into a train, validation, and test set to ensure an independent evaluation. While these splits should be different from one another to test the generalization capabilities of a method, the data splits should also have similar characteristics to ensure a fair evaluation.

To address this conflict of objectives, we developed a method to find a Pareto-optimal solution for this multi-objective optimization problem (MOOP). For this purpose, the optimization uses the NSGA-II~\cite{Deb2002} genetic algorithm with random sampling, polynomial mutation~\cite{Deb2007}, and simulated binary crossover~\cite{Deb2007} without duplicates. The optimization problem

\begin{equation}
    \begin{aligned}
        \min_{s \in S}(f_{1}(s), f_{2}(s), ..., f_{12}(s)) \\
        |{S_{\text{train}}}| = 0.7 |S|, \enskip |{S_{\text{test}}}| = 0.2 |S|
    \end{aligned}
    \label{eq:moop}
\end{equation}

is given as a minimization problem with constraints, where $S$ is the set of all scenes $s$ defined by their scene properties (number of object classes, scene tags, sample timestamps, and ego positions) with cardinality $|S|$. The objectives $f_{1}(s)\,\text{to}\,f_{8}(s)$ are the minimization of the deviations of the discrete distributions of annotations across dataset splits, which are chosen to be the class distribution and the distributions of the seven individual scene tag categories. The objectives $f_{9}(s)\,\text{and}\,f_{10}(s)$ aim to maximize the intra-split standard deviation of temporal (sample timestamps) and spatial (ego vehicle positions) components of the scenes. Finally, $f_{11}(s)\,\text{and}\,f_{12}(s)$ aim to maximize the inter-split Kullback–Leibler (KL) divergence of the temporal and spatial components. The split sizes are set to \SI{70}{\percent}, \SI{10}{\percent}, and \SI{20}{\percent} of all scenes for the train, validation, and test splits.

\subsection{Privacy}
\label{sec:privacy}
Compliance with data protection measures is a top priority, which is why the whole dataset is anonymized. The anonymization includes not only the blurring of faces and license plates in the image data but also the anonymization of timestamp information. Nevertheless, the temporal consistency is still guaranteed for all samples and sensor data. As shown by Alibeigi~et~al.~\cite{Alibeigi2023}, the chosen image anonymization should not affect the downstream perception tasks.

\section{Tasks}
\label{sec:tasks}
\MANDataset{} supports a multitude of different perception tasks through its multimodal nature, sequential data structure, and rich annotations. These tasks include object detection, tracking, prediction, and localization, even if we want to put special emphasis on 3D object detection.

The detection task requires detecting 3D bounding boxes of 12 different object classes which are a subset of the original 27 annotation classes. These classes are selected based on the nuScenes~\cite{Caesar2020} detection task and the insides gained during our labeling campaign. As a result, the evaluation excludes object classes that are not present in all data splits and combines subclasses with high inter-class confusion (seen during labeling) in accordance with our hierarchical class structure.

The evaluation is based on the nuScenes Detection Score (NDS), which is a weighted sum of the mean Average Precision (mAP) and five True Positive (TP) metrics~\cite{Caesar2020}. In contrast to an Intersection over Union (IoU) based mAP, the NDS uses a distance-based mAP and averages the individual Average Precision (AP) values not only over the classes but also over four discrete distance thresholds. In addition to the mAP, the NDS takes five TP metrics into account, which are the Average Translation Error (ATE), Average Scale Error (ASE), Average Orientation Error (AOE), Average Velocity Error (AVE), and Average Attribute Error (AAE). Further details on the NDS and its calculation can be found in~\cite{Caesar2020}.

In contrast to the NDS, our detection range is not limited to \SI{50}{\metre} during evaluation but considers objects of up to \SI{150}{\metre} around the vehicle. This should emphasize the development of long-range detection methods, which are crucial for the safe operation of autonomous vehicles on highways. Furthermore, we introduce the animal and traffic sign classes as two additional detection classes, leading to 12 distinct detection classes.

% Aspects
% \begin{itemize}
%     \item Multitask (detection, tracking, prediction, localization)
%     \item Comparability
%     \item Relevanz
%     \item Class aggregation (combination of multiple classes to one)
% \end{itemize}

\section{Experiments}
\label{sec:results}
We provide baseline 3D object detection results for all three sensor modalities, which are shown in Table~\ref{tab:results}. The camera results are based on the PETR~\cite{Liu2022} model architecture with a pre-trained FCOS3D (V-99-eSE) backbone and an adjusted detection head to support 12-class classification. The detection distance is set to $\pm \SI{150}{\metre}$ and the final image resolution is chosen to be $320\times800$ pixels. The radar baseline is set by RadarGNN~\cite{Fent2023}, which was trained on fused, ego-motion compensated 4D radar point clouds with 6 aggregated sweeps. The node features consisted of the rcs values, timestamps, connectivity degrees, and the relative radial velocities of the radar points. Lastly, a CenterPoint~\cite{Yin2021} model was trained on fused ego-motion compensated lidar point clouds with 3 aggregated sweeps cropped to a $\SI{300}{\metre}\times\SI{300}{\metre}$ grid with a voxel resolution of $(\SI{0.1}{\metre}, \SI{0.1}{\metre}, \SI{0.2}{\metre})$. The model used seven detection heads and was evaluated for 12 detection classes, according to our metric definition.

The results suggest that the detection quality of diverse object classes (like 'animal' or 'other vehicle') is insufficient and that the overall detection quality decreases significantly with increasing distance. Moreover, the results show a reduced detection quality in winter conditions and tunnels. Furthermore, it can be observed that the camera model provides insufficient results for minority classes and struggles with challenging weather and lighting conditions, whereas the radar-based detection model provides insufficient results for all non-metallic object classes. Moreover, it can be observed that the radar performance is negatively affected by reflections within tunnels or container terminals. The lidar performance, on the other hand, is reduced under rainy or foggy conditions but also struggles within tunnel scenarios and seems to be more sensitive to the number of available training samples. In general, the results show that a more robust long-range perception is required for safe autonomous trucking.

\begin{table}
  \caption{Baseline results on the test split of the \MANDataset{} dataset v1.0.}
  \label{tab:results}
  \centering
  \begin{tabular}{lllccccccc}
    \toprule
     Method & Dataset & Modality & mAP & NDS & ATE & ASE & AOE & AVE & AAE \\
    \midrule
    PETR~\cite{Liu2022} & MAN & camera & 0.02 & 0.12 & 1.13 & 0.69 & 0.65 & 1.50 & 0.56 \\
    RadarGNN~\cite{Fent2023} & MAN & radar & 0.07 & 0.11 & 0.89 & 0.81 & 1.13 & 8.00 & 0.57 \\
    CenterPoint~\cite{Yin2021} & MAN & lidar & 0.27 & 0.41 & 0.41 & 0.35 & 0.28 & 2.73 & 0.20 \\
    \bottomrule
  \end{tabular}
\end{table}

\section{Conclusion}
\label{sec:summary}
In this work, we presented \MANDataset{}, the first dataset for autonomous trucking. It is a large-scale multimodal dataset for the development of autonomous truck perception applications. The dataset consists a diverse set of scenes recorded in a multitude of different environmental conditions and provides \SI{360}{\degree} coverage of all sensor modalities, including 4D radar data. It contains carefully reviewed 3D bounding boxes, tracking information, and rich annotations for all objects within more than \SI{230}{\metre} range. Furthermore, we introduced a generic method for compiling meaningful and balanced data splits. We provide baseline results for 3D object detection and promote research in all areas of perception, tracking, and prediction. For this purpose, we will maintain a public leaderboard and provide annotation instructions, taxonomy specifications, and a development kit. With this dataset, we aim to accelerate the development of autonomous trucking.

\section{Limitations}
\label{sec:limits}
The dataset is limited to measurements on public roads and logistic terminals in Germany recorded on a single autonomous test truck. The data distribution represents the operational design domain of a long-haul truck and is not representative for other distribution haulage. The dataset is manually annotated and, therefore, subject to human annotation errors even if our extensive quality assurance process seeks to minimize them. The extrinsic sensor calibration can be affected by cabin movements, the IMUs are exposed to vehicle vibrations, and the ego vehicle position is subject to GNSS inaccuracies even with RTK correction.

\section{Societal Impact}
\label{sec:society}
Despite the potential benefits of autonomous trucks the societal impact of the technology must be considered. Since autonomous trucks operate on public roads, their safety is a major concern that must be assured and cannot be compromised for economic reasons. While safety assurance is not just the key deployment factor, it is also the main driver for public acceptance and trust~\cite{Engstrm2018}. Therefore, regulatory authorities have to establish standards that ensure the safety of autonomous trucking and allow cross-border operation. Besides that, the impact on labor must be taken into account. While some studies expect job losses through autonomous vehicles~\cite{Engstrm2018}, others argue that trained personnel will still be needed in the overall logistics process to supervise the system or carry out manual tasks~\cite{Escherle2023.01}. Recent studies show that advanced automation can even help to improve working conditions and therefore, counteract the shortage of drivers and minimize turnover rates~\cite{Escherle2023.02}. Ultimately, regulatory measures must be put in place to ensure that the interests of the general public are protected.

\section*{Acknowledgment}
\label{sec:thanks}
\MANDataset{} would not have been possible without the support of many. For this reason, we would like to thank the whole MAN team for their dedicated and passionate work. We would like to thank the MAN workshop and integration team for their tireless support and the MAN management and project team for the strong promotion of this project. Last but not least, we want to thank our TRATON and Scania colleagues for their friendly support and expertise. The dataset annotation and quality assurance were handled by b-plus. This work was partially funded by the government-supported ATLAS-L4 project with grand no. 19A21048I.
% \newpage

{
\small
\bibliographystyle{abbrvnat}
\bibliography{main}
}

% Appendix
\newpage
\appendix

\addcontentsline{toc}{section}{Appendices}
\newcommand{\hbAppendixPrefix}{A}
\renewcommand{\thefigure}{\hbAppendixPrefix\arabic{figure}}
\setcounter{figure}{0}
\renewcommand{\thetable}{\hbAppendixPrefix\arabic{table}} 
\setcounter{table}{0}

% rotated table
\newcommand{\STAB}[1]{\begin{tabular}{@{}c@{}}#1\end{tabular}}

\section{Appendix}
% Optionally include supplemental material (complete proofs, additional experiments and plots) in appendix.
% All such materials \textbf{SHOULD be included in the main submission.}

The appendix provides additional details on the \MANDataset{} dataset and the conducted experiments. The sections are structured in accordance with the main body.

\subsection{Sensor Synchronization}
% \begin{itemize}
%     \item Synchronization (plot)
%     \item Synchronization quality metrics
% \end{itemize}

\begin{figure}[h]
    \centering
    \begin{subfigure}{\linewidth}
      \centering
      \includegraphics[width=13.97cm,trim={0 0.4cm 0 0.4cm},clip]{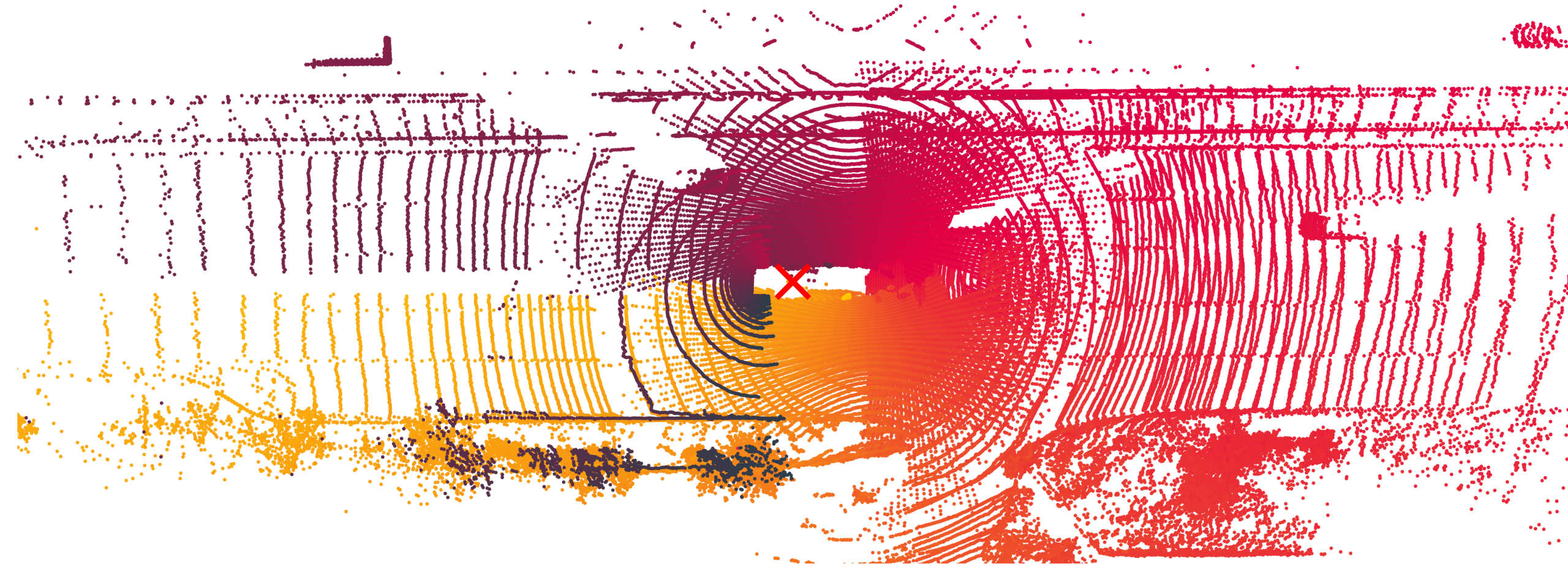}
    \end{subfigure}
    \hfill
    \begin{subfigure}{\linewidth}
      \centering
      \includegraphics[width=10.03cm]{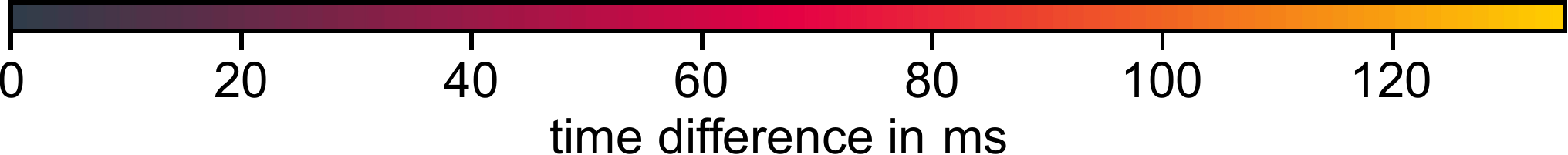}
    \end{subfigure}
    \caption{Lidar point clouds of all six lidar sensors color-coded by their time difference to the earliest recorded point in all six point clouds.}
    \label{fig:sync}
\end{figure}

Sensor synchronization is especially important to ensure temporal consistency in multimodal datasets. As mentioned in Section~\ref{sec:sync}, all cabin-mounted lidar sensors are phase-synchronized, such that their lidar points are temporally consistent in a clockwise manner. However, this approach leads to a temporal discontinuity at one point of a full cycle. This point is chosen to be in the back of the vehicle where the rear lidar is located. Furthermore, perfect consistency cannot be achieved for the top-mounted lidar sensors due to their tilted axis of rotation and constant rotational speed. Nevertheless, local deviations are small, such that the fused point clouds of a single sample have similar temporal properties as a single top-mounted lidar sensor. This temporal consistency of the lidar point clouds is shown in Figure~\ref{fig:sync}. 

\begin{figure}[h]
    \centering
    \includegraphics[width=13.97cm]{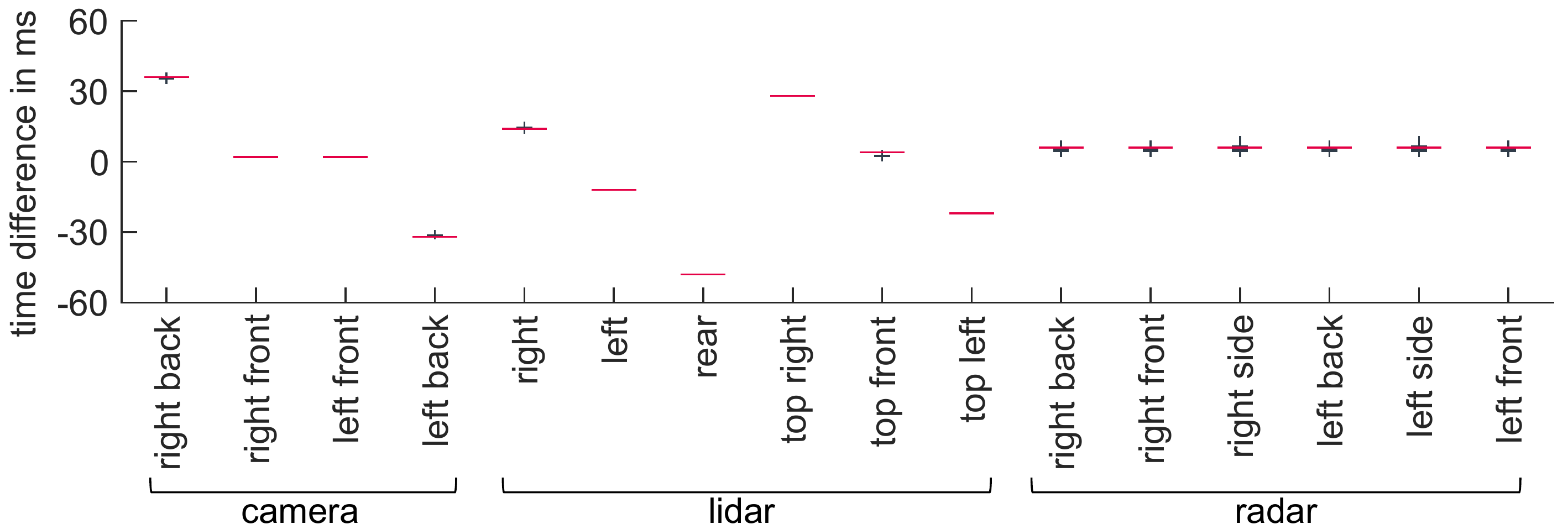}
    \caption{Differences of the sensor timestamps to the sample timestamp represented by their median offset (red lines), interquartile range (blue boxes), and whiskers (blue lines) across all samples.}
    \label{fig:offset}
\end{figure}

To comply with the nuScenes data format all annotations are made at a specific point in time sampled at a frequency of \SI{2}{\hertz}. The annotation time is chosen to be the point in time when all lidar scans align at the longitudinal axis in front of the vehicle. Therefore, the individual sensor timestamps have an offset to the sample (annotation) timestamp, as shown in Figure~\ref{fig:offset}. To account for the differences in the recording time, the annotations are based on a fused and ego-motion compensated point cloud, as described in Section~\ref{sec:annotation}. However, this procedure can only compensate for ego-motion and not for target-motion. This can lead to inconsistencies, especially at the point of temporal discontinuity in the back of the vehicle.

\subsection{Sensor Calibration}
% \begin{itemize}
%     \item Calibration projection lidar onto camera
%     \item Calibration quality metrics
% \end{itemize}

\begin{figure}[h]
    \centering
    \begin{subfigure}{0.495\linewidth}
      \centering
      \includegraphics[width=6.9cm]{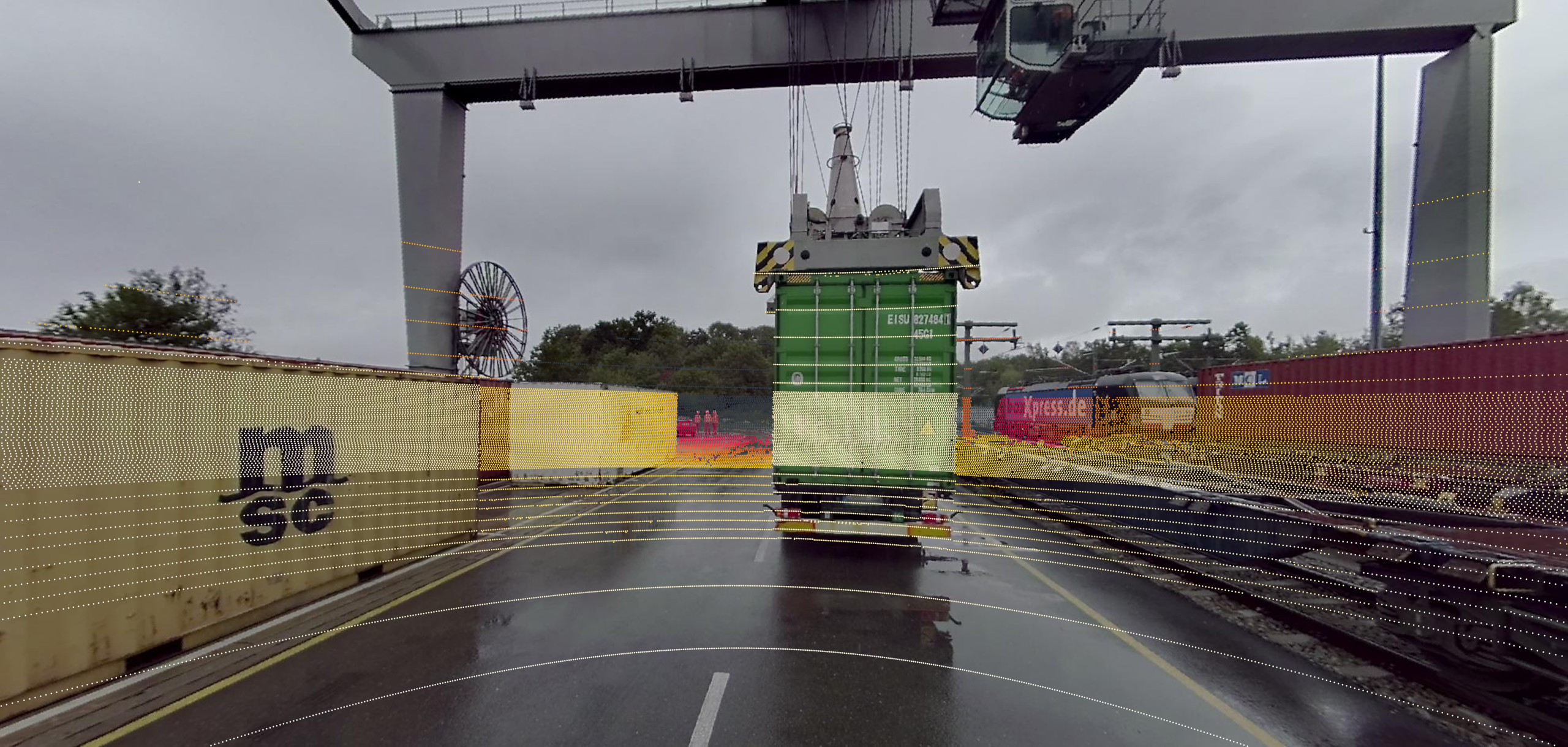}
    \end{subfigure}
    \hfill
    \begin{subfigure}{0.495\linewidth}
      \centering
      \includegraphics[width=6.9cm]{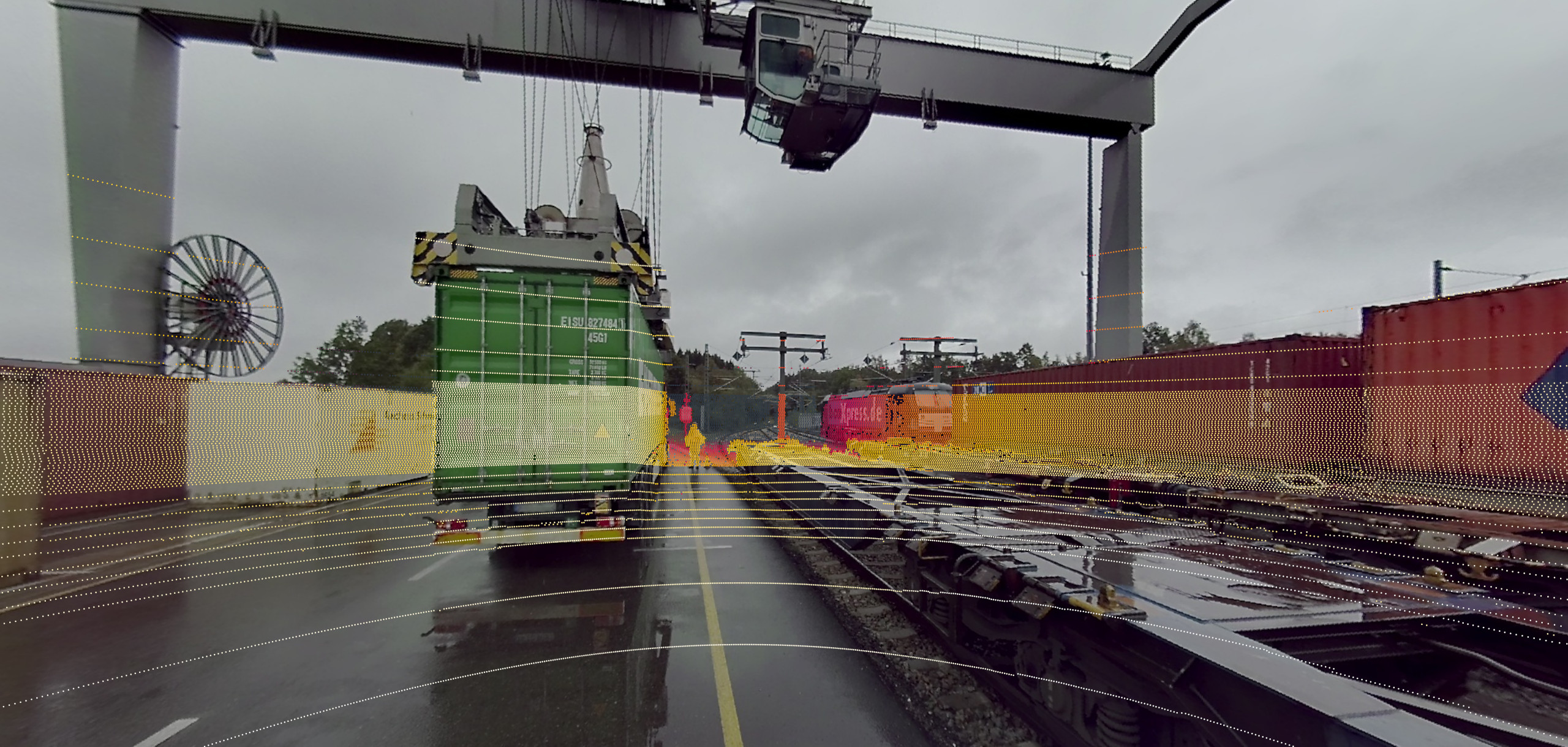}
    \end{subfigure}

    \caption{Projection of the left lidar sensor onto the front left camera image (left) and the right lidar sensor on the right front camera image (right). The color of the points indicates the relative distance.}
    \label{fig:cal}
\end{figure}

The quality of the sensor calibration is illustrated by a projection of the lidar points onto the camera images, as shown in Figure~\ref{fig:cal}. It is important to note that a matching projection can only be achieved within a static scenario for multiple reasons. First, differences in the recording time caused by small deviations in the sensor synchronization (Figure~\ref{fig:offset}) can lead to changes in the captured environment. Second, the rotating measurement principle of the lidar sensors leads to a temporal gradient (from left to right) in the recording of the lidar points (Figure~\ref{fig:sync}). Third, the rolling shutter effect of the cameras causes a temporal gradient (from top to bottom) in the recording of the pixel rows. All these effects can lead to inconsistencies in the measurements of the different sensor modalities.

\subsection{Sensor Data}
% \begin{itemize}
%     \item fields (radar, lidar, IMU, GPS)
%     \item Camera undistortion (distorted vs. undistorted)
%     \item Data quality metrics (see test scrips)
% \end{itemize}

\begin{figure}[h]
    \centering
    \begin{subfigure}{0.43\linewidth}
      \centering
      \includegraphics[height=3.72cm]{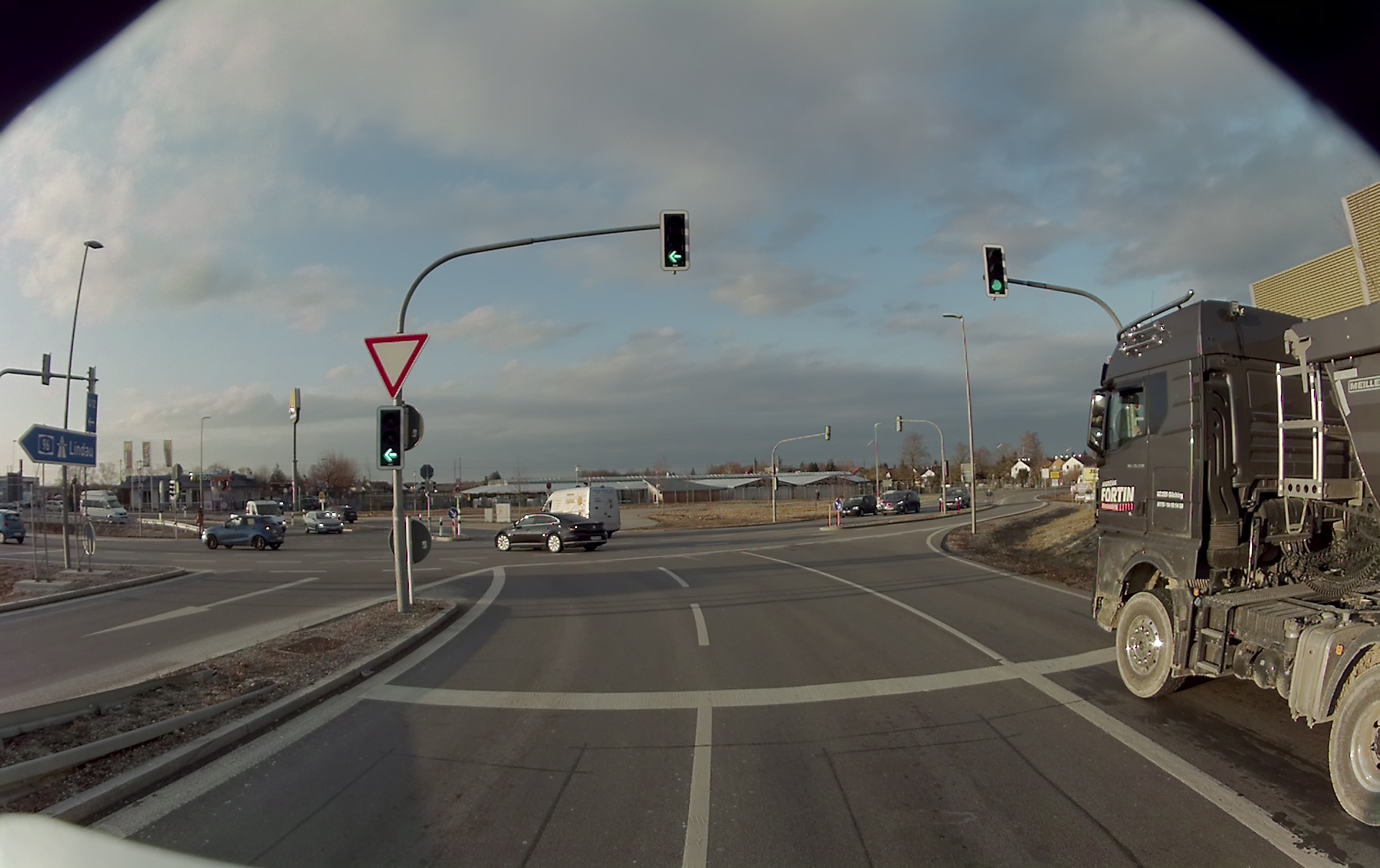}
    \end{subfigure}
    \hfill
    \begin{subfigure}{0.56\linewidth}
      \centering
      \includegraphics[height=3.72cm]{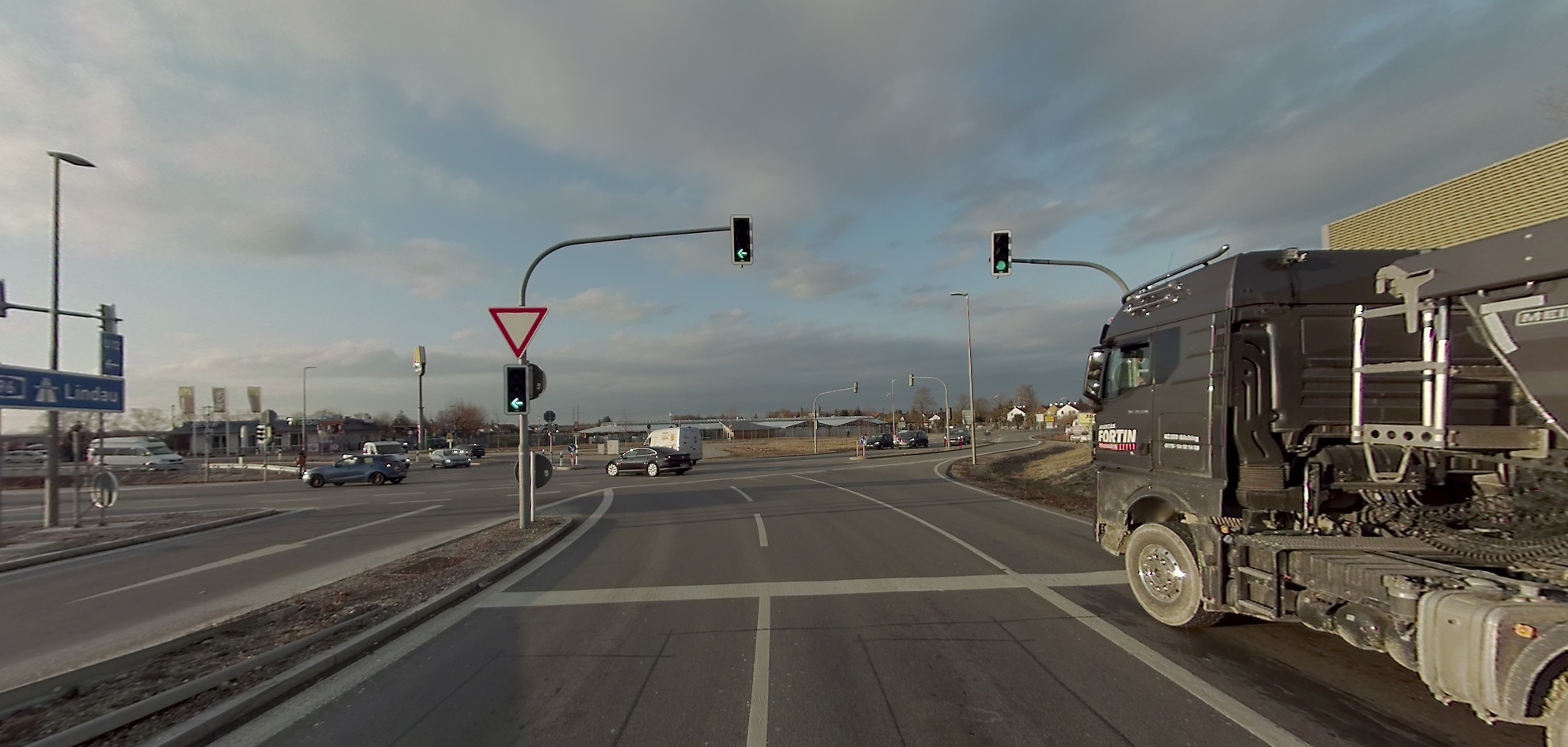}
    \end{subfigure}

    \caption{Comparison of the original (left) and the undistorted (right) camera image.}
    \label{fig:undistortion}
\end{figure}

The provided sensor data is subject to multiple processing steps described in Section~\ref{sec:data}. Within this process, the camera data is mapped to a pinhole camera model to ensure its compatibility with the nuScenes data format and most existing perception methods. Therefore, the images of the wide-angle cameras are undistorted and cropped to remove black spots from the optics. The differences between the original and processed camera images can be seen in Figure~\ref{fig:undistortion}.

\subsection{Scene Selection}
% \begin{itemize}
%     \item Geographical coverage (map)
%     \item Class distribution (per split)
% \end{itemize}

\begin{figure}[ht]
    \centering
    \includegraphics[width=13.97cm]{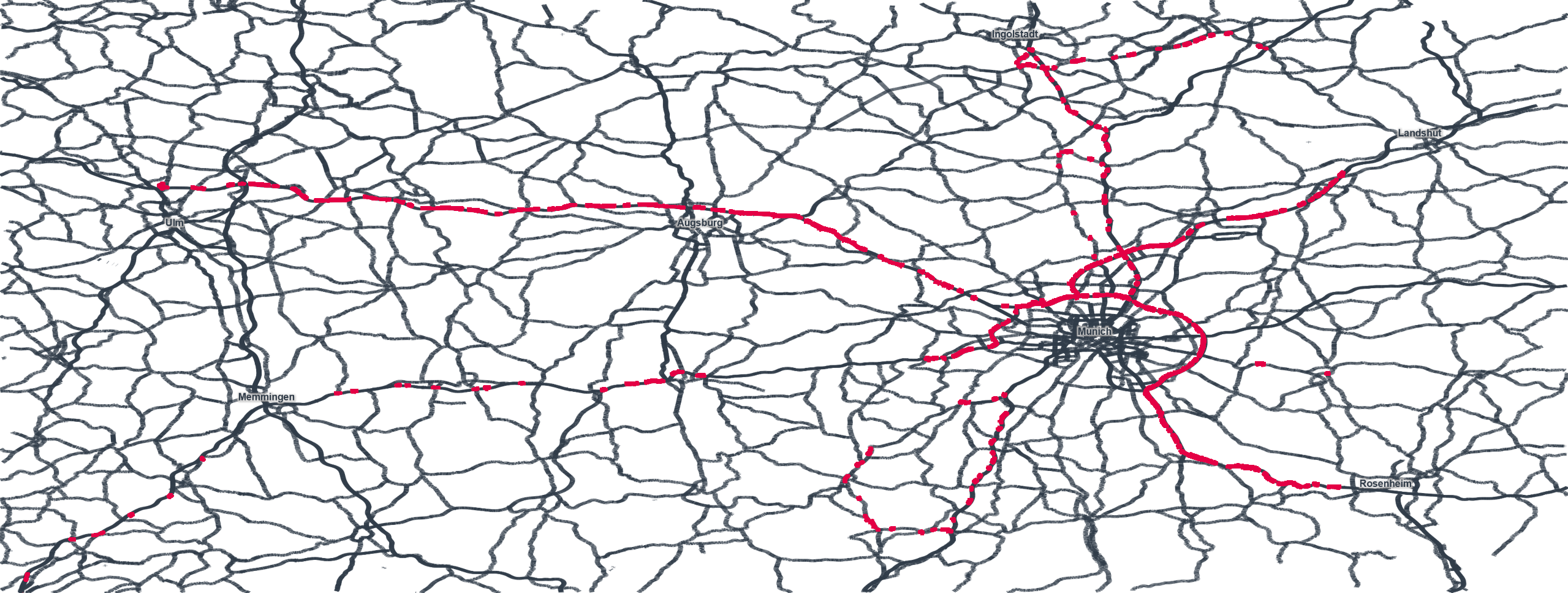}
    \caption{Georaphical locations of all recorded scenes represented by their ego-poses shown in red.}
    \label{fig:map}
\end{figure}

The scene selection aims to select diverse scenarios representative of a long-haul truck's operation, as described in Section~\ref{sec:selection}. This selection primarily consists of highway, terminal, and rural scenes as well as scenarios from city and residential areas. To still be able to provide a diverse set of scenes, \MANDataset{} covers a geographical area of \SI{100}{\kilo\metre\squared}, measured by the union area of the \SI{150}{\metre}-diluted ego-poses, as defined in \cite{Sun2020}. The geographical locations of all included scenes are shown in Figure~\ref{fig:map}. While \MANDataset{} does not provide map data, the included GNSS data enables the usage of open geographic databases (like OpenStreetMap) to utilize external map data, as shown in Figure~\ref{fig:map}.

\subsection{Data Annotation}
% \begin{itemize}
%     \item Attribute distribution
%     \item Tracking length per object class
%     \item Spatial annotation distribution (heatmap)
%     \item Annotation distribution (range, per sample)
%     \item Annotation distribution (velocity)
%     \item Number of different categories per sample
%     \item Points (lidar, radar) per object over range
%     \item Annotation quality measures (labeling)
% \end{itemize}

\begin{figure}[h]
  \centering
  \includegraphics[width=13.97cm]{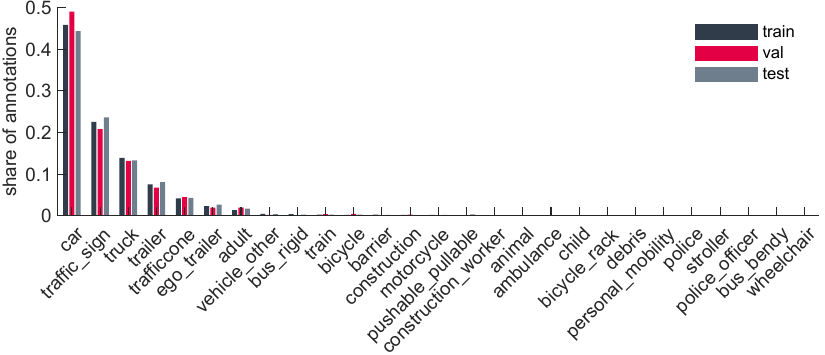}
  \caption{Distribution of the object categories grouped by dataset split.}
  \label{fig:splits}
\end{figure}

This section provides additional information on the included annotations of the dataset. As mentioned in Section~\ref{sec:annotation}, the annotations comprise scene tags, 3D bounding boxes, tracking IDs, and object attributes. While the distribution of the scene tags and the distribution of the bounding box categories can be seen in Figure~\ref{fig:tags} and Figure~\ref{fig:classes}, respectively, Figure~\ref{fig:splits} shows the distribution of the object categories by dataset split. It can be seen that the different splits show a similar class distribution with slight variations due to the optimization criteria on temporal and spatial divergence.

\begin{figure}[ht]
    \centering
    \begin{subfigure}{.3\linewidth}
        \centering
        \includegraphics[width=4.191cm,trim={0 0.8cm 0 0},clip]{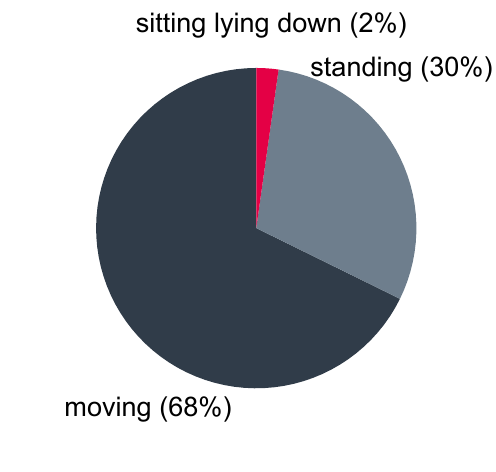}
        \caption{Pedestrian}
    \end{subfigure}
        \hfill
    \begin{subfigure}{.3\linewidth}
        \centering
        \includegraphics[width=4.191cm,trim={0 0.8cm 0 0},clip]{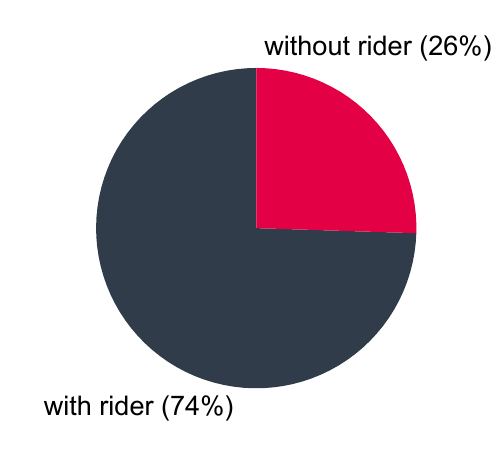}
        \caption{Cycle}
    \end{subfigure}
       \hfill
    \begin{subfigure}{.3\linewidth}
        \centering
        \includegraphics[width=4.191cm,trim={0 0.8cm 0 0},clip]{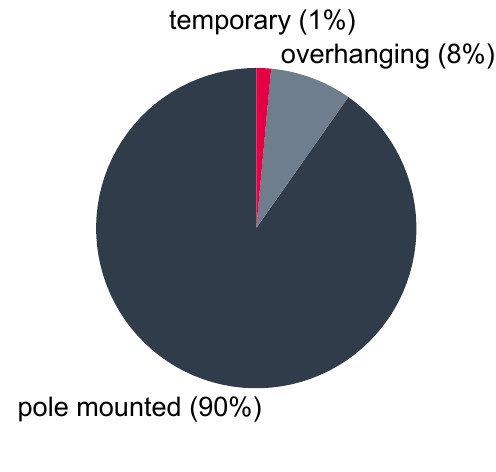}
        \caption{Traffic Sign}
    \end{subfigure}
    
    \bigskip
    \begin{subfigure}{.15\linewidth}
      \centering
    \end{subfigure}
    \hfill
    \begin{subfigure}{.3\linewidth}
      \centering
      \includegraphics[width=4.191cm,trim={0 0.8cm 0 0},clip]{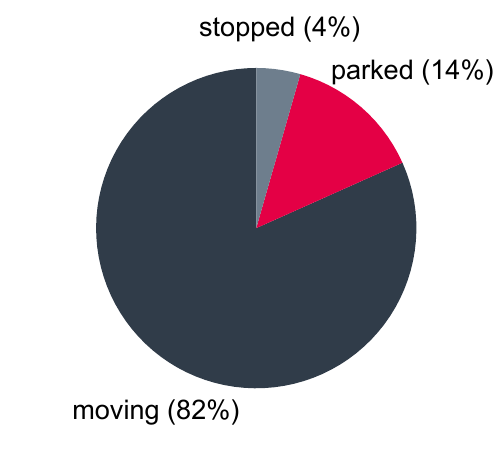}
      \caption{Vehicle}
    \end{subfigure}
    \hfill
    \begin{subfigure}{0.3\linewidth}
      \centering
      \includegraphics[width=4.191cm,trim={0 0.8cm 0 0},clip]{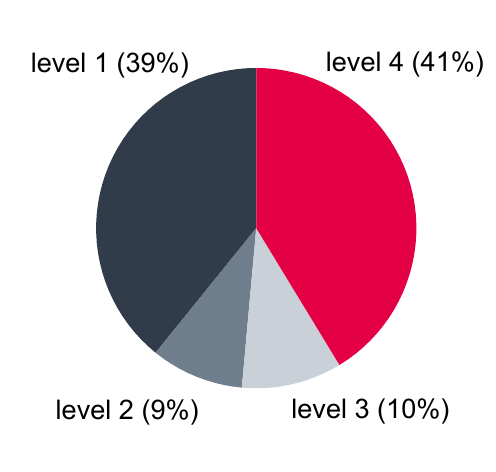}
      \caption{Visibility}
    \end{subfigure}
    \hfill
    \begin{subfigure}{.15\linewidth}
      \centering
    \end{subfigure}

    \caption{Distribution of the 15 different object attributes shown by the five attribute groups.}
    \label{fig:attributes}
\end{figure}

In addition to the analysis of category and tag distribution, Figure~\ref{fig:attributes} provides an analysis of the distribution of the different object attributes. It can be seen that most objects are moving and characterized by high visibility. Still, \SI{39}{\percent} of the objects have a visibility of less than \SI{41}{\percent} (visibility level 1). This number reflects the challenges of long-range perception and occlusions by large vehicles.

\begin{figure}[ht]
  \centering
  \includegraphics[width=13.97cm]{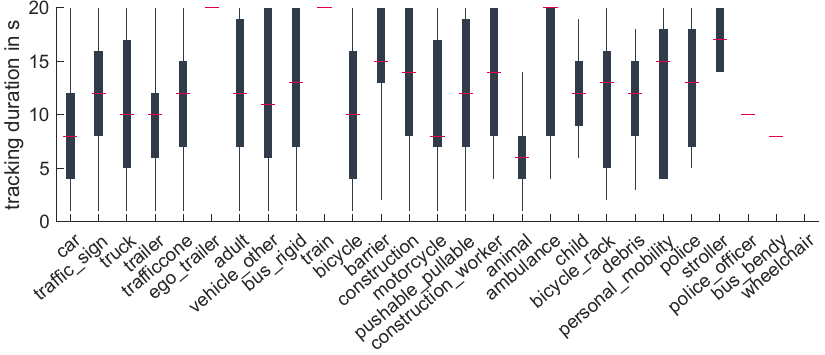}
  \caption{Instance tracking duration for all instances of the dataset represented by their median tracking duration (red lines), interquartile range (blue boxes), and whiskers (blue lines).}
  \label{fig:tracks}
\end{figure}

Another important measure for tracking and prediction is the tracking duration of different objects, which is why Figure~\ref{fig:tracks} is included. The analysis shows that the mean tracking duration across all categories is \SI{9.4}{\second}, with a median duration of \SI{8}{\second} for the car class. Since established prediction metrics, like the nuScenes or Argoverse2 metric, use a prediction horizon of \SI{6}{\second}, the majority of all tracks are suitable for this purpose. It can also be seen that the ego-trailer class is reliably tracked throughout the whole scene (if a trailer is attached), which provides the possibility for developing trailer detection or articulation angle estimation methods.

\begin{figure}[ht]
    \centering
    \includegraphics[width=13.97cm]{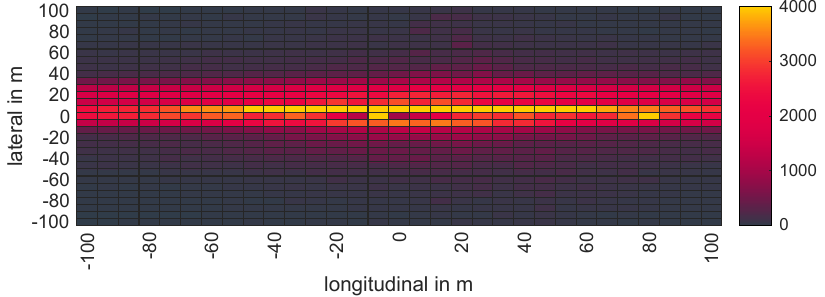}
    \caption{Spatial distribution of the annotated objects in the vehicle coordinate frame. The color indicates the number of annotated objects within the particular cell throughout the overall dataset.}
    \label{fig:heatmap}
\end{figure}

In addition to the class and attribute distribution, we also analyzed the spatial distribution of the annotated objects. Figure~\ref{fig:heatmap} shows that the majority of the objects occur in the front or the back of the vehicle and that most of the objects are to the left of the ego vehicle. This is probably because a significant proportion of the scenes were recorded on highways, and (in Germany) vehicles must always use the rightmost lane whenever possible. Therefore, most of the other vehicles are either part of the oncoming traffic or overtaking our ego vehicle, which results in the shown distribution. 

\begin{figure}[h]
    \centering
    \begin{subfigure}{0.49\linewidth}
      \centering
      \includegraphics[width=6.84cm]{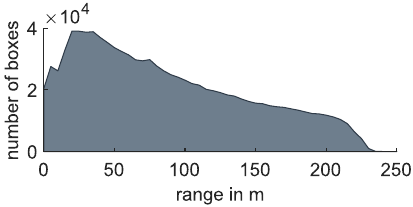}
    \end{subfigure}
    \hfill
    \begin{subfigure}{0.49\linewidth}
      \centering
      \includegraphics[width=6.84cm]{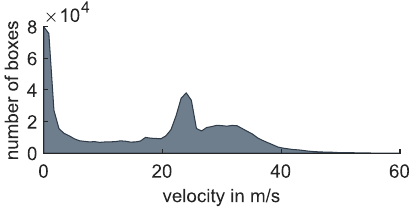}
    \end{subfigure}

    \caption{Number of annotated 3D bounding boxes by range (right) and by velocity (left). Note that the velocity distribution only considers moving objects with more than \SI[per-mode=symbol]{0.5}{\metre\per\second}.}
    \label{fig:boxes}
\end{figure}

Another unique property of \MANDataset{} is the provision of long-range detections and annotations of objects with high velocities. As listed in Table~\ref{tab:comparison}, the 99.9th percentile of all annotated bounding box distances is \SI{226}{\metre} and \SI{50}{\percent} of all annotations are beyond \SI{75}{\metre}. This number is significantly higher than the \SI{14}{\percent} of the Argoverse2 dataset, whereas nuScenes and Waymo have less than \SI{1}{\percent} objects beyond \SI{75}{\metre}~\cite{Wilson2021}. A detailed distribution of the \MANDataset{} annotation distances can be seen in Figure~\ref{fig:boxes}.

In addition, \MANDataset{} includes a significant amount of objects with high absolute velocities. The average velocity of all objects within the dataset is \SI[per-mode=symbol]{14}{\metre\per\second} and \SI{40}{\percent} of all objects are moving faster than \SI[per-mode=symbol]{20}{\metre\per\second}. The velocity distribution of all moving objects (objects with a velocity higher than \SI[per-mode=symbol]{0.5}{\metre\per\second}) is shown in Figure~\ref{fig:boxes}. It can be seen that a significant amount of objects are moving faster than \SI[per-mode=symbol]{30}{\metre\per\second}, while a local maximum can be observed at \SI[per-mode=symbol]{24}{\metre\per\second}. This can be explained by the fact that, in Germany, the maximum permitted speed for vehicles heavier than \SI{3.5}{\tonne} is \SI[per-mode=symbol]{80}{\kilo\metre\per\hour} (\SI[per-mode=symbol]{22}{\metre\per\second}).

\begin{figure}[h]
    \centering
    \begin{subfigure}{0.49\linewidth}
      \centering
      \includegraphics[width=6.84cm]{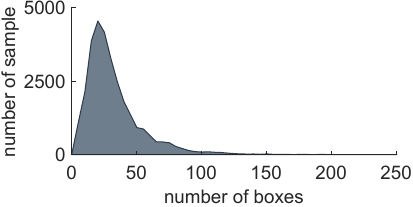}
    \end{subfigure}
    \hfill
    \begin{subfigure}{0.49\linewidth}
      \centering
      \includegraphics[width=6.84cm]{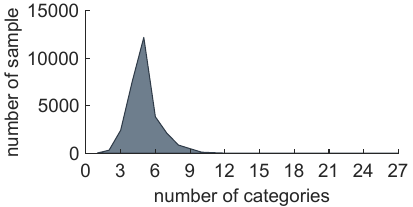}
    \end{subfigure}

    \caption{Number of dataset samples over the number of annotated 3D bounding boxes (right) and over the number of different object categories (left).}
    \label{fig:sample}
\end{figure}

The \MANDataset{} dataset provides an average of 34 objects per sample, which is comparable to the 33 objects per sample of the nuScenes dataset but less than the Argoverse2 or Waymo dataset with 75 and 61, respectively~\cite{Wilson2021}. This is probably because the aforementioned datasets were recorded in city environments, while \MANDataset{} was mainly recorded on highways and terminal environments. However, the validity of such a comparison is limited due to the different taxonomies of the datasets. Nevertheless, the sample-based object distributions of the \MANDataset{} dataset are shown in Figure~\ref{fig:sample}.

% \subsection{Tasks}
% \begin{itemize}
%     \item Class aggregation
%     \item Detection range per class
%     \item Annotation confusion matrix
% \end{itemize}

\subsection{Experiments}
% \begin{itemize}
%     \item Per class results
%     \item Results over range bins
%     \item Per tag results (area, daytime, lighting, season, weather)
%     \item Ressources, training time, inference time
% \end{itemize}

Experiments were conducted for all three sensor modalities. The camera results are based on the PETR~\cite{Liu2022} architecture, the radar baseline on RadarGNN~\cite{Fent2023}, and the lidar results on CenterPoint~\cite{Yin2021}. To be specific, the camera baseline utilizes a PETR~\cite{Liu2022} model with a pre-trained FCOS3D (V-99-eSE) backbone with two multi-scale feature map outputs. Subsequently, a Feature Pyramid Network (FPN) neck is used for feature alignment and a transformer-based detection head is used with six decoder layers. The 900 query points are initialized within an area of $\SI{320}{\metre}\times\SI{320}{\metre}\times\SI{20}{\metre}$ and both the features as well as queries are positional encoded with a sine encoding. The training is Non Maximum Suppression (NMS) free and uses a Hungarian Assignment method instead, combined with a Focal Loss. The implementation is based on the MMDetection3D framework and uses common image augmentation techniques like random rotation, scaling, or translation. The final model is trained on an Nvidia A100 GPU for 48 epochs (\SI{38}{\hour}) and a batch size of 8.

The radar model uses the translation invariant RadarGNN~\cite{Fent2023} variant with a KNN-based graph construction method with a maximum of 20 neighbors in a radius of \SI{1}{\metre} and directed edges with associated features. The main body of the model consists of five consecutive Message-Passing Neural Network (MPNN) layers and a separate classification and regression head. The training is based on a Cross-Entropy and Huber Loss function and does not utilize any data augmentation methods. The final training is conducted on an Nvidia A100 GPU with a batch size of 32 for 30 epochs.

The lidar method is based on a CenterPoint~\cite{Yin2021} model with a voxel-based encoder based on the SECOND architecture, a FPN neck, a Deep \& Cross Network (DCN) CenterHead, and circular NMS. The training scheme uses different data augmentation methods, including random global rotation, translation, scaling, flipping, point shuffling, and object sampling. The overall model is based on the MMDetection3D framework and was trained for 20 epochs on two Nvidia A100 GPUs for \SI{181}{\hour} with a batch size of~16 per GPU.

\begin{figure}[t]
    \centering
    \begin{subfigure}{0.325\linewidth}
      \centering
      \includegraphics[width=4.54cm]{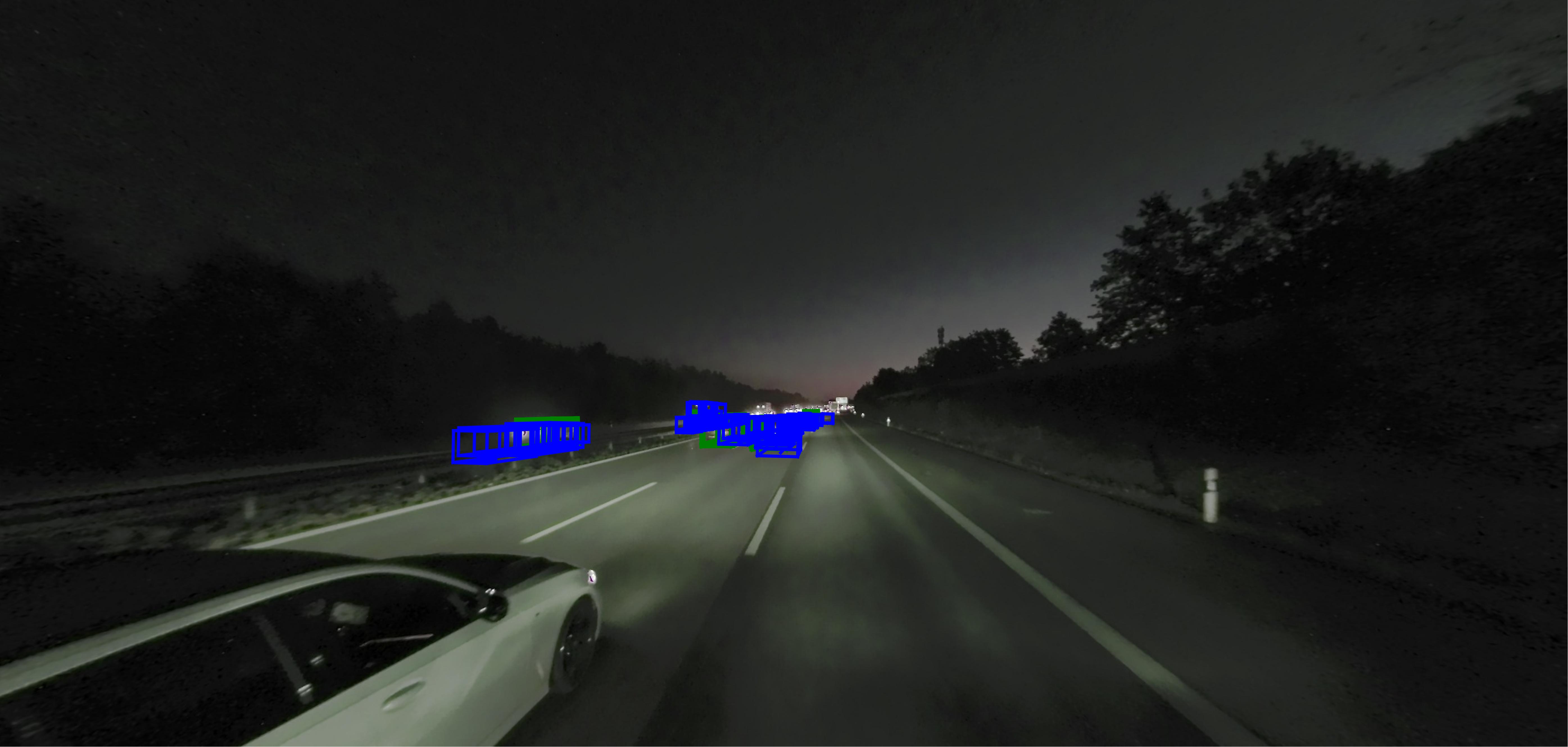}
    \end{subfigure}
    \hfill
    \begin{subfigure}{0.325\linewidth}
      \centering
      \includegraphics[width=4.54cm]{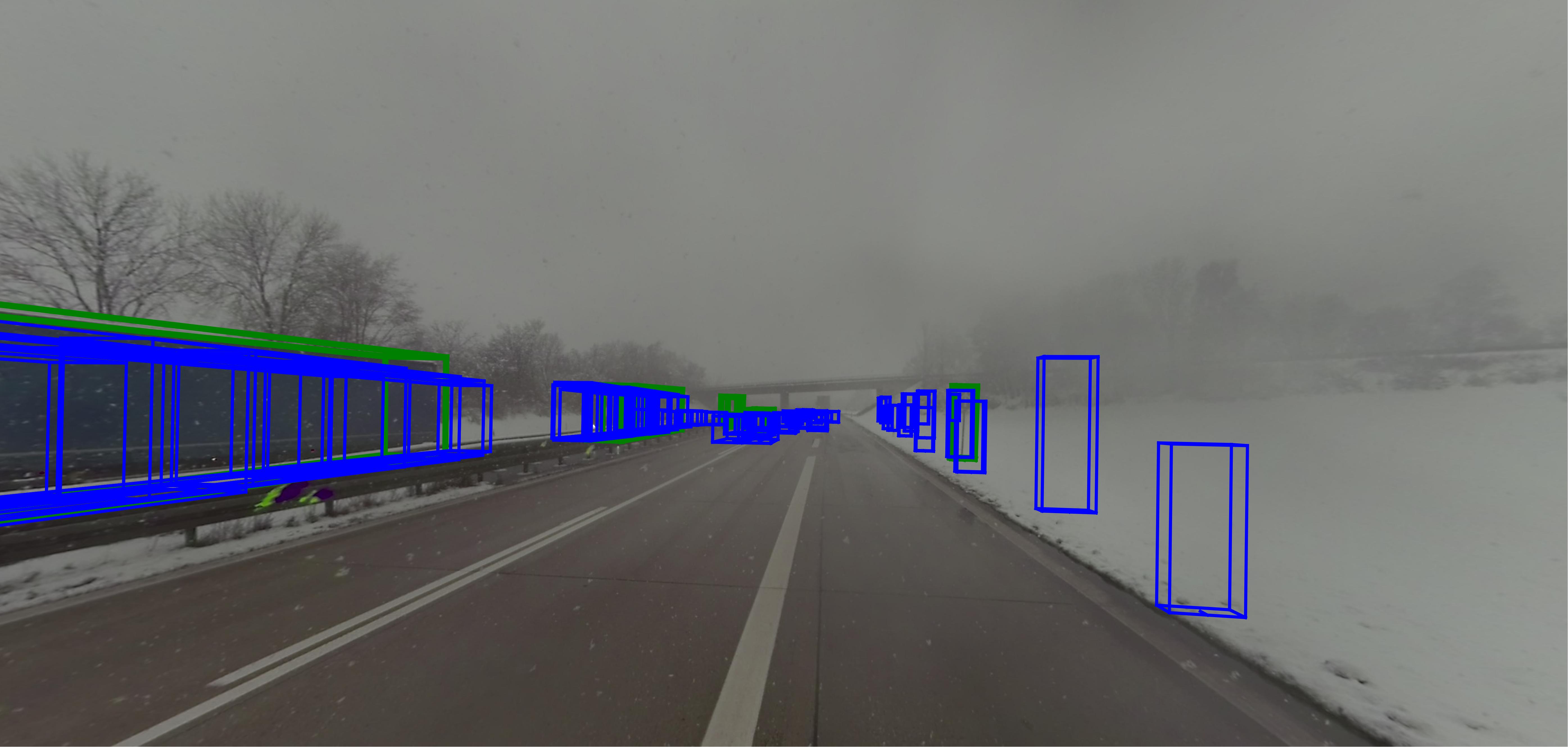}
    \end{subfigure}
    \hfill
    \begin{subfigure}{0.325\linewidth}
      \centering
      \includegraphics[width=4.54cm]{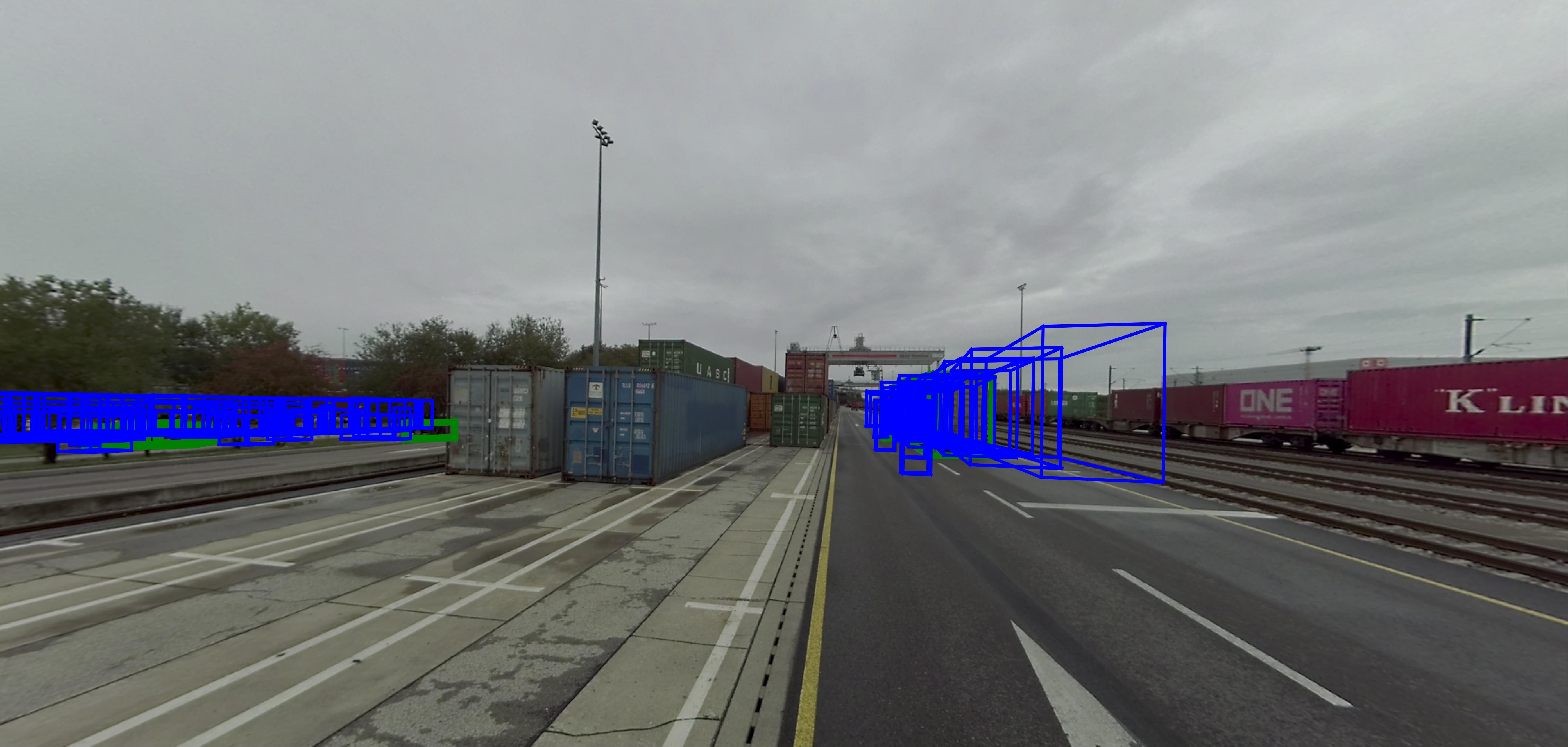}
    \end{subfigure}
    \\ \vspace{8pt}
    \begin{subfigure}{0.325\linewidth}
      \centering
      \includegraphics[height=4.54cm,angle=-90,origin=c]{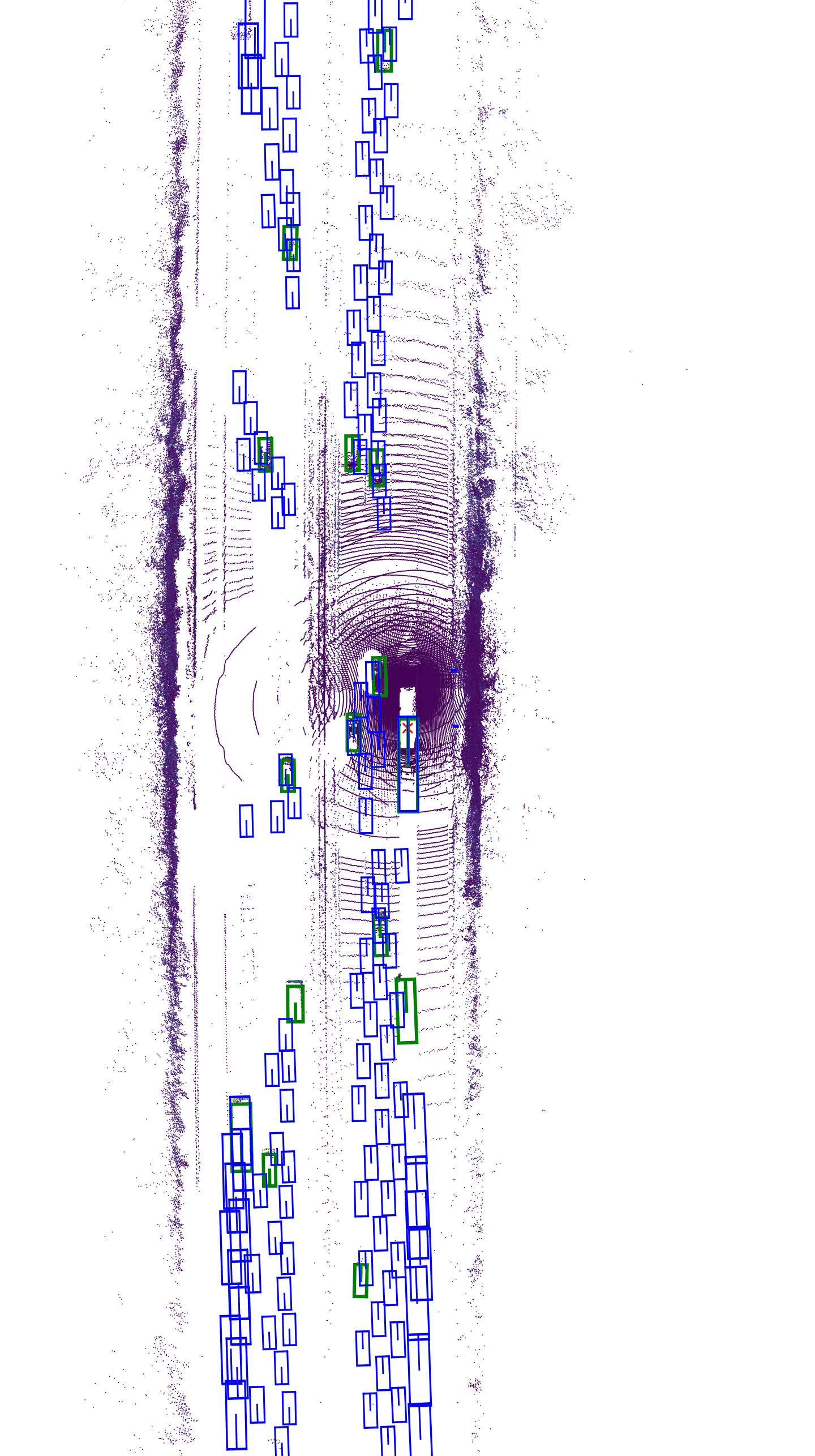}\vspace{-24pt}
      \caption{night}
    \end{subfigure}
    \hfill
    \begin{subfigure}{0.325\linewidth}
      \centering
      \includegraphics[height=4.54cm,angle=-90,origin=c]{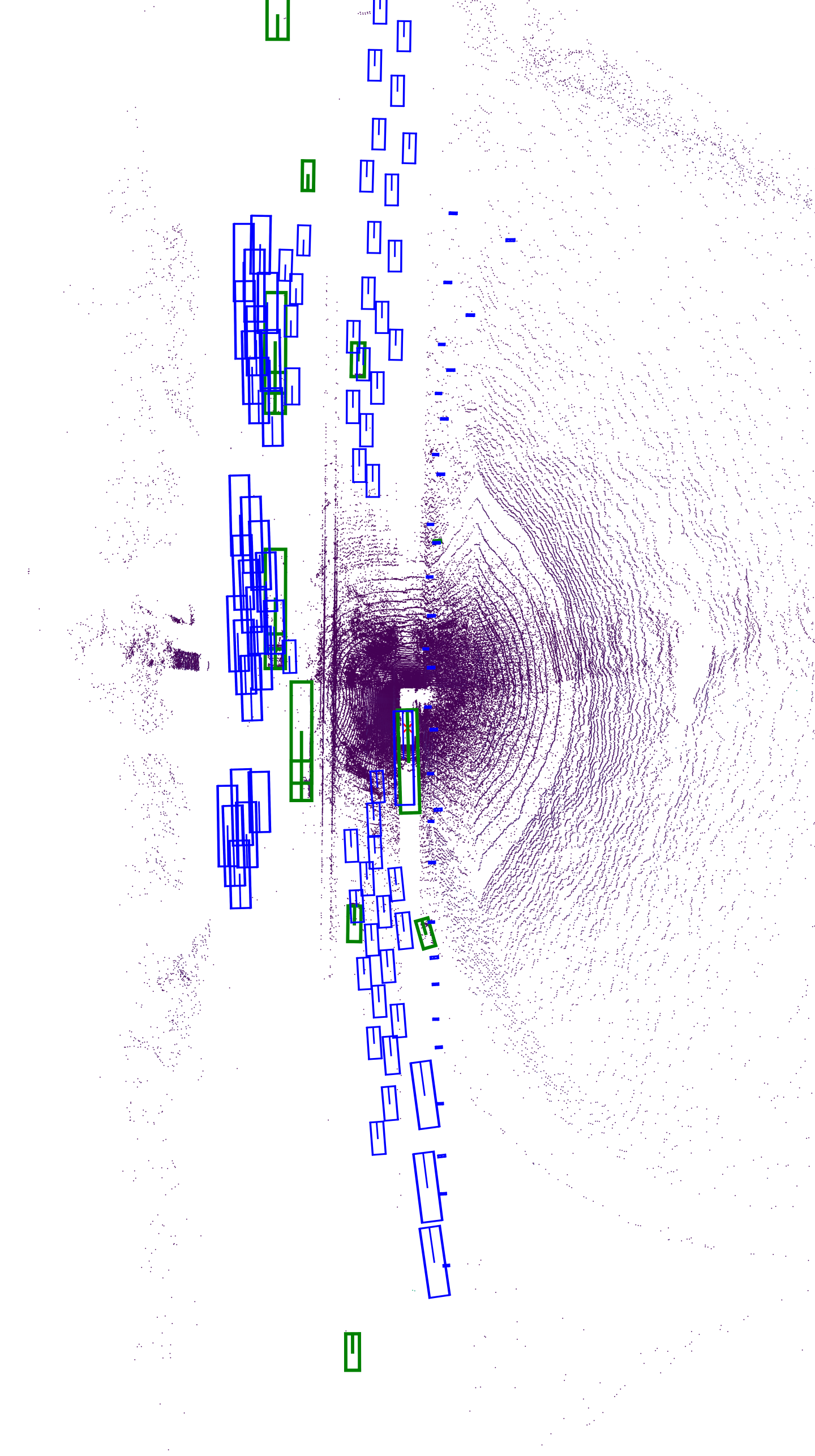}\vspace{-24pt}
      \caption{snow}
    \end{subfigure}
    \hfill
    \begin{subfigure}{0.325\linewidth}
      \centering
      \includegraphics[height=4.54cm,angle=-90,origin=c]{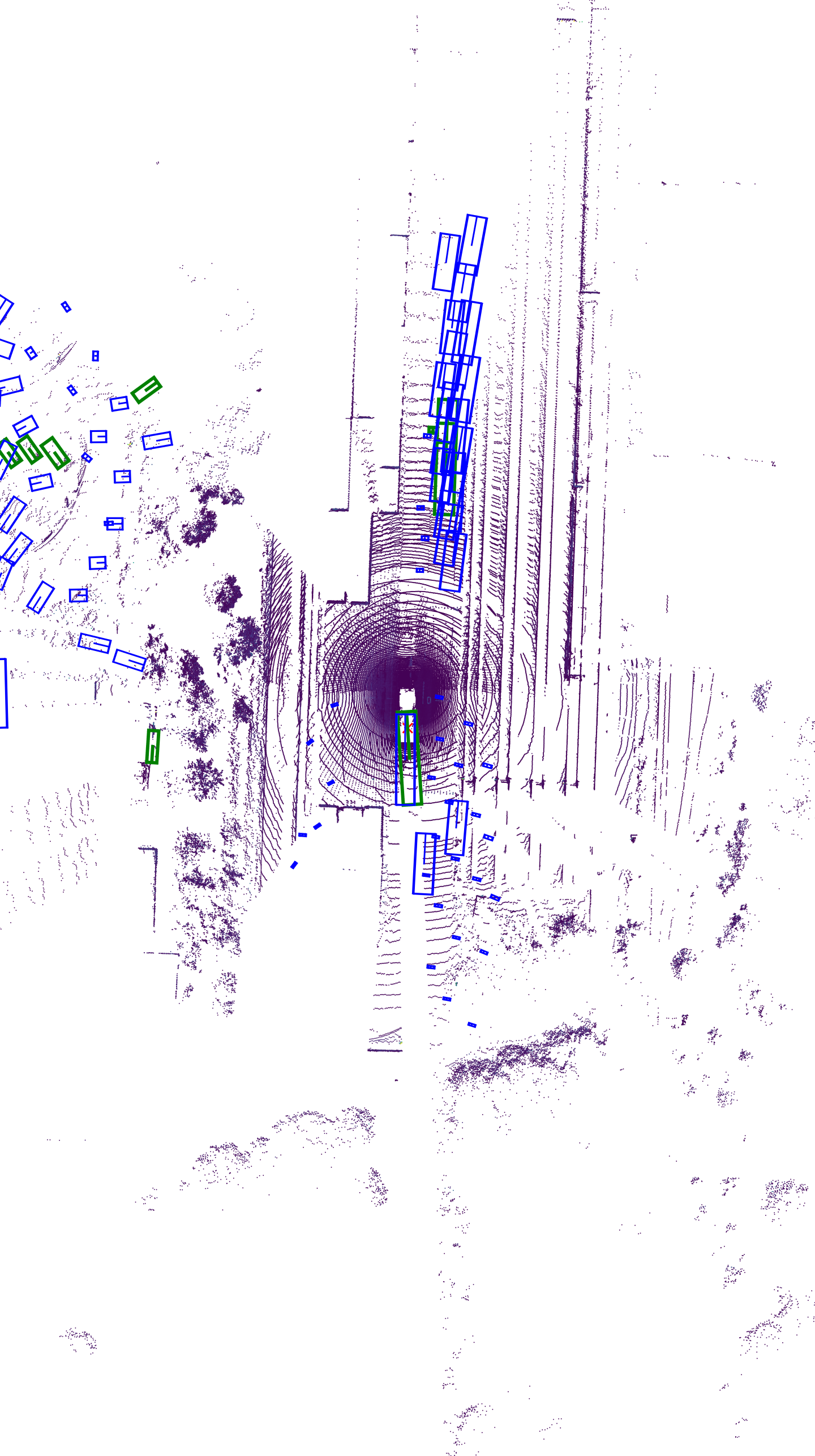}\vspace{-24pt}
      \caption{terminal}
    \end{subfigure}

    \caption{Example visualizations of the PETR model on the validation split of the \MANDataset{} dataset v1.0. The front left camera is shown on the top and the fused lidar point cloud on the bottom. Model predictions are shown in \textcolor{blue}{blue} and the ground truth in \textcolor{green}{green}.}
    \label{fig:camera_examples}
\end{figure}

\begin{figure}[t]
    \centering
    \begin{subfigure}{0.325\linewidth}
      \centering
      \includegraphics[width=4.54cm]{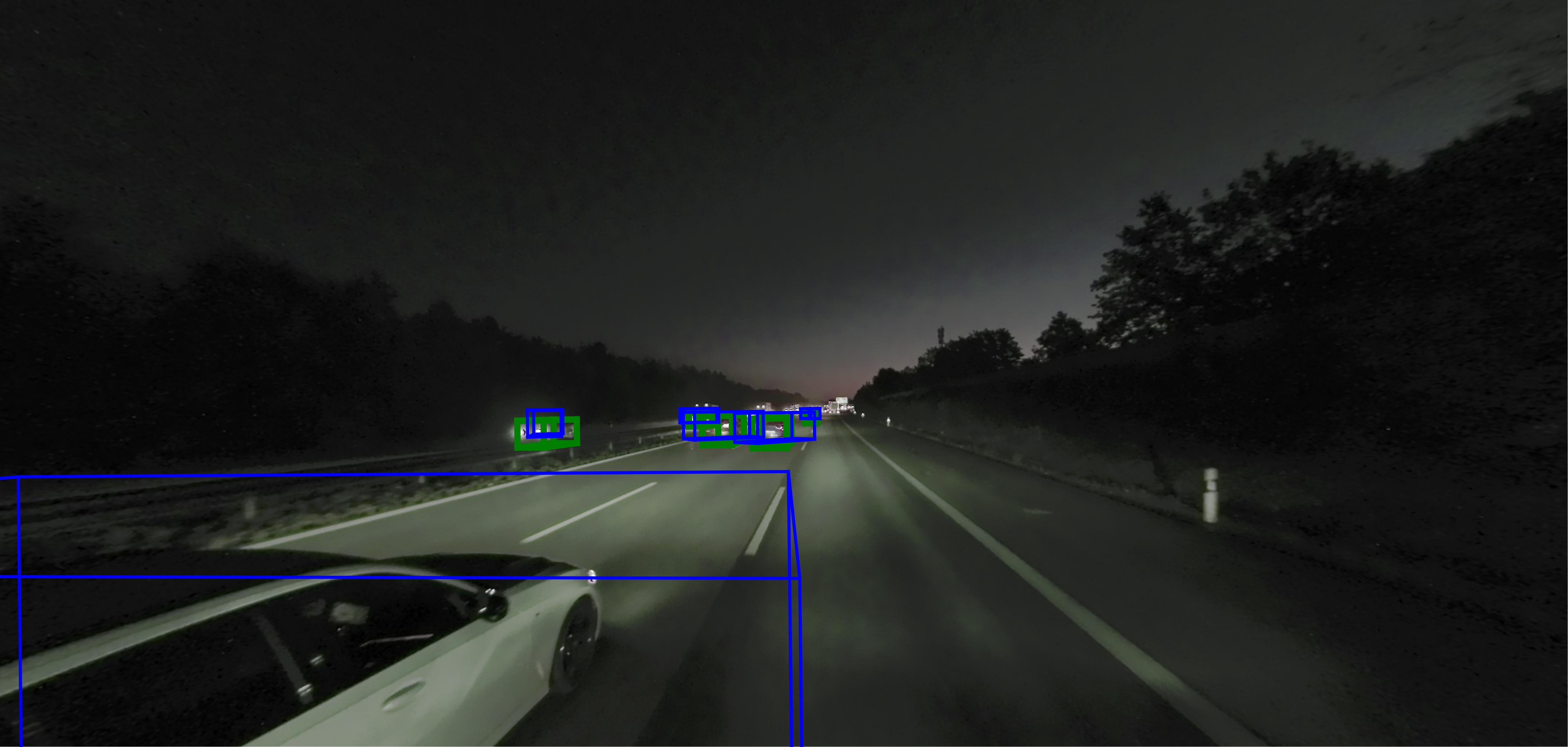}
    \end{subfigure}
    \hfill
    \begin{subfigure}{0.325\linewidth}
      \centering
      \includegraphics[width=4.54cm]{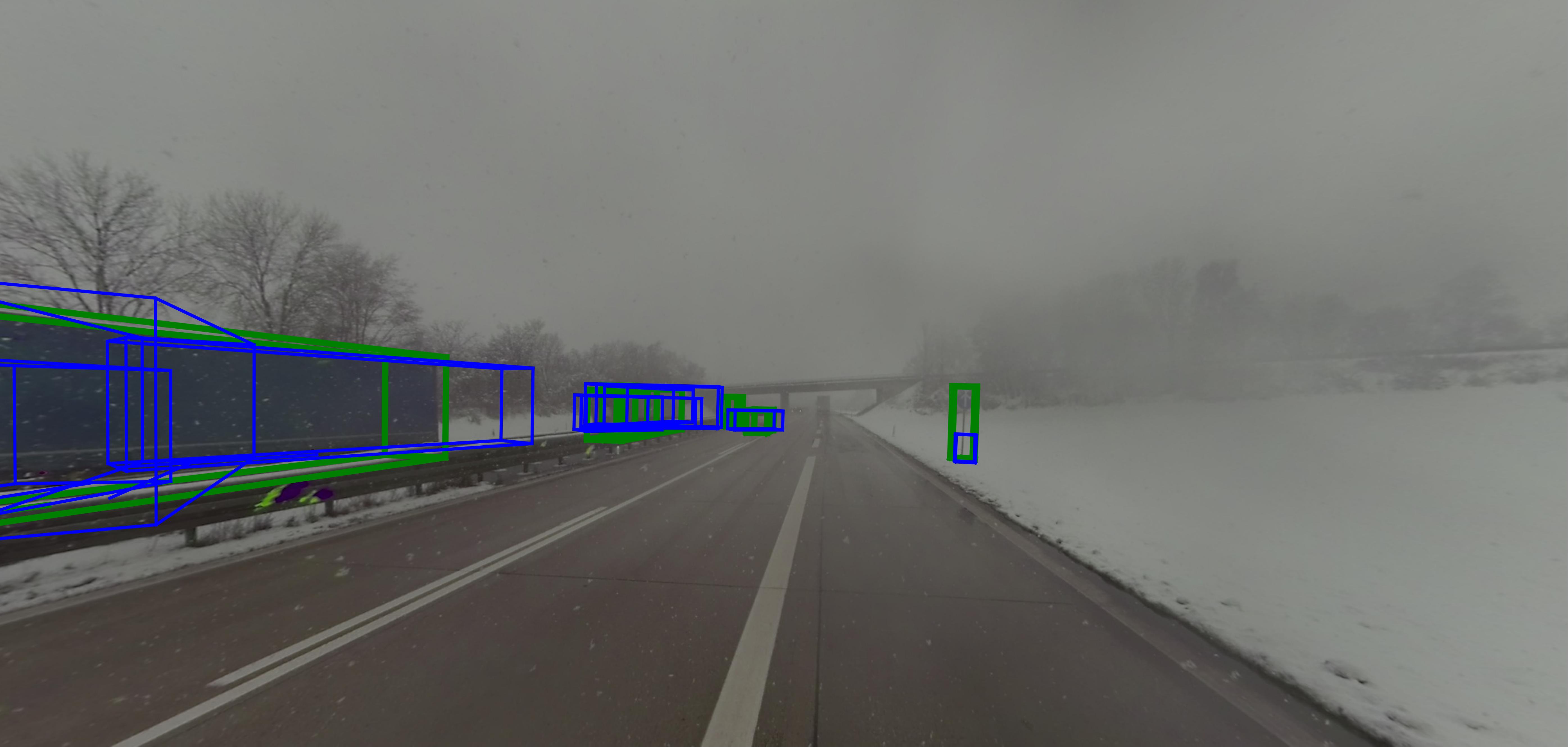}
    \end{subfigure}
    \hfill
    \begin{subfigure}{0.325\linewidth}
      \centering
      \includegraphics[width=4.54cm]{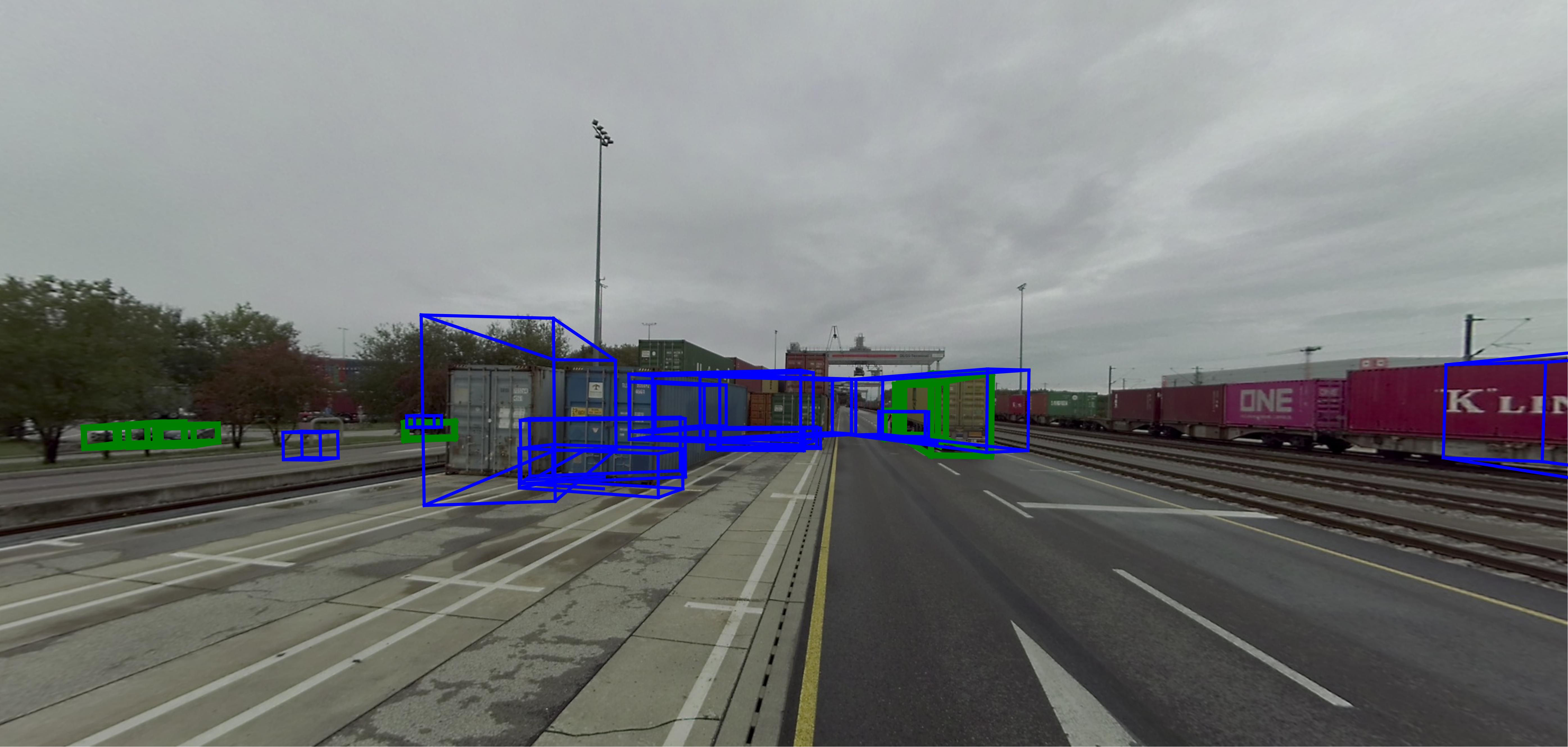}
    \end{subfigure}
    \\ \vspace{8pt}
    \begin{subfigure}{0.325\linewidth}
      \centering
      \includegraphics[height=4.54cm,angle=-90,origin=c]{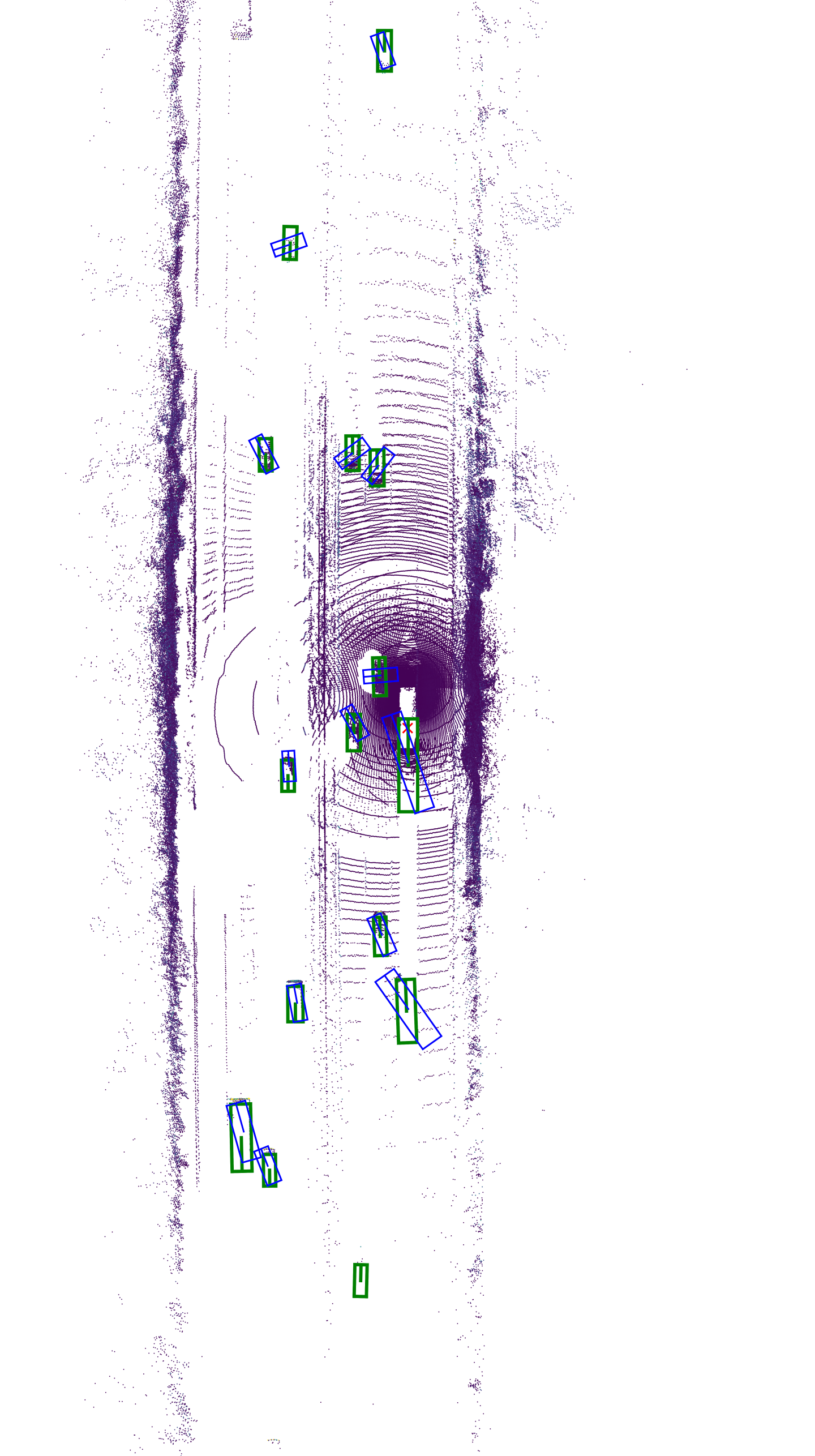}\vspace{-24pt}
      \caption{night}
    \end{subfigure}
    \hfill
    \begin{subfigure}{0.325\linewidth}
      \centering
      \includegraphics[height=4.54cm,angle=-90,origin=c]{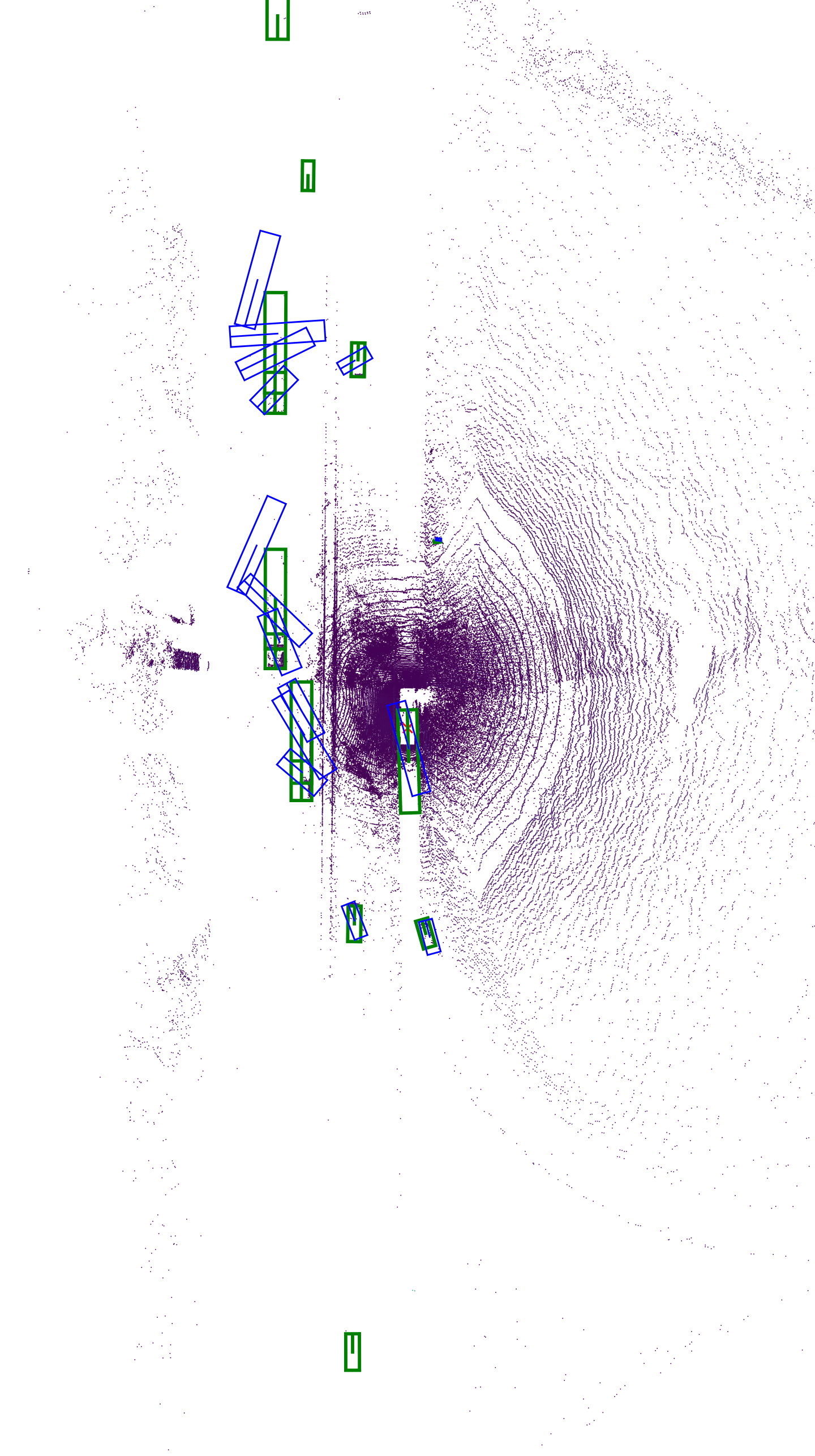}\vspace{-24pt}
      \caption{snow}
    \end{subfigure}
    \hfill
    \begin{subfigure}{0.325\linewidth}
      \centering
      \includegraphics[height=4.54cm,angle=-90,origin=c]{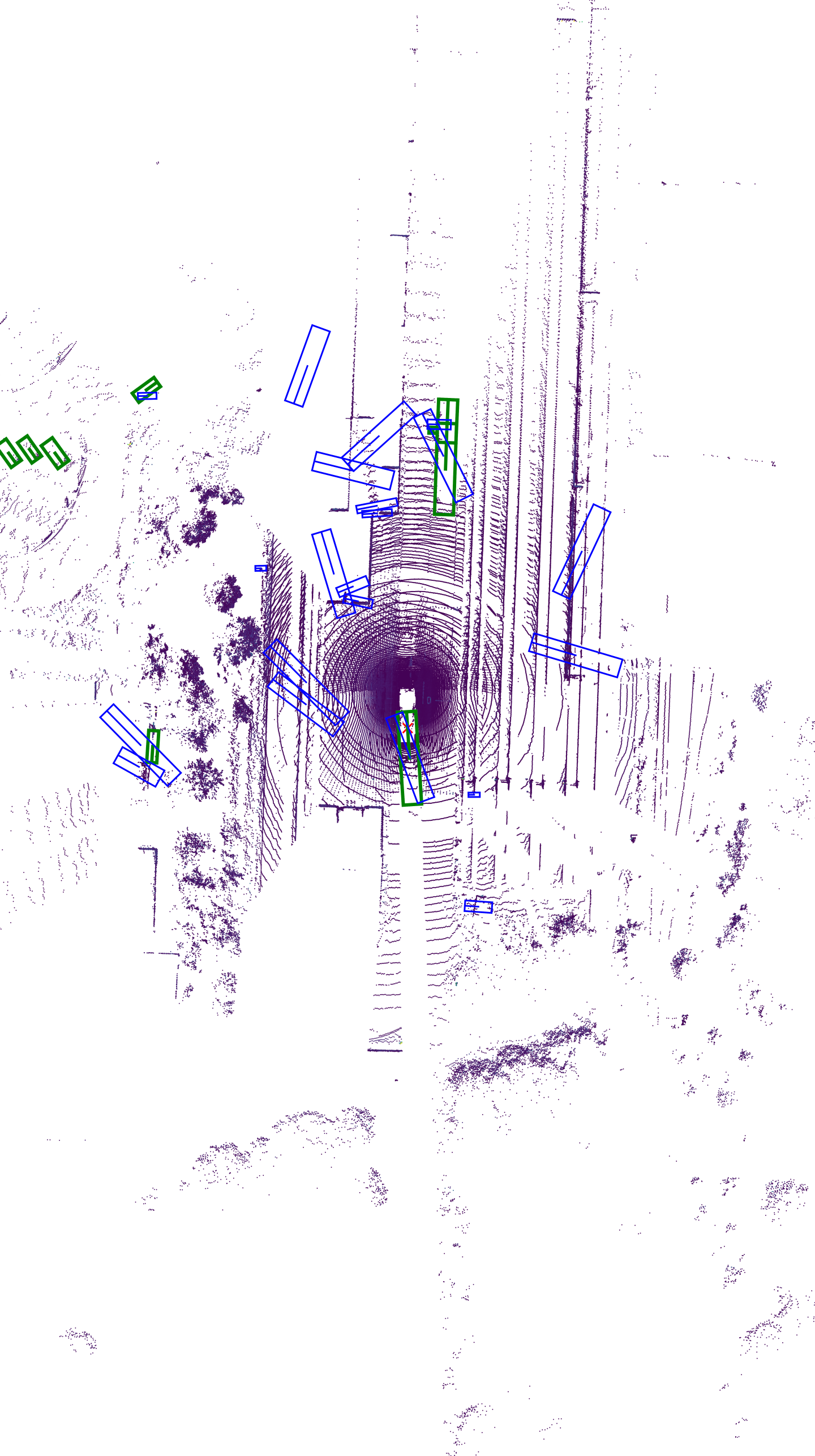}\vspace{-24pt}
      \caption{terminal}
    \end{subfigure}

    \caption{Example visualizations of the RadarGNN model on the validation split of the \MANDataset{} dataset v1.0. The front left camera is shown on the top and the fused lidar point cloud on the bottom. Model predictions are shown in \textcolor{blue}{blue} and the ground truth in \textcolor{green}{green}.}
    \label{fig:radar_examples}
\end{figure}

\begin{figure}[t]
    \centering
    \begin{subfigure}{0.325\linewidth}
      \centering
      \includegraphics[width=4.54cm]{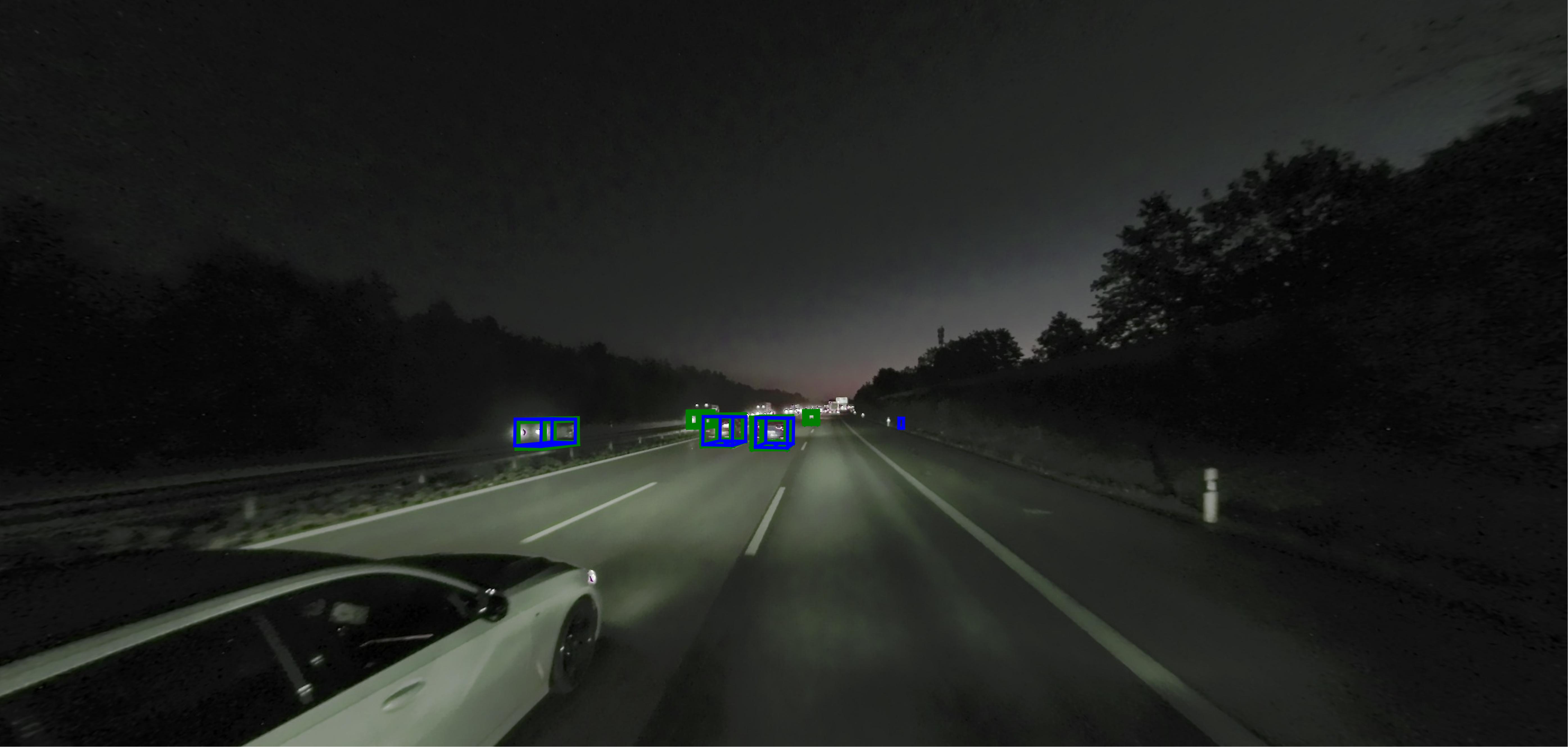}
    \end{subfigure}
    \hfill
    \begin{subfigure}{0.325\linewidth}
      \centering
      \includegraphics[width=4.54cm]{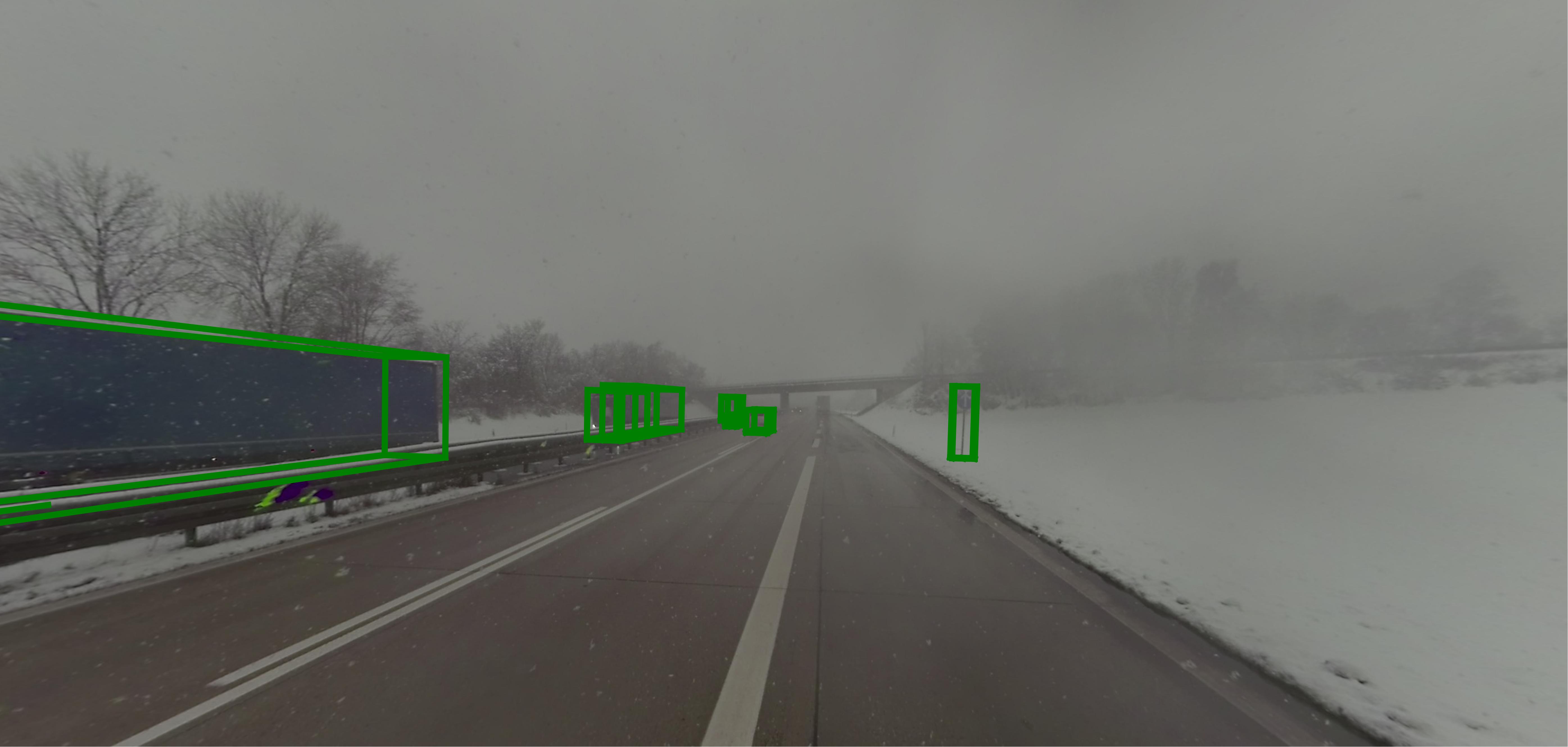}
    \end{subfigure}
    \hfill
    \begin{subfigure}{0.325\linewidth}
      \centering
      \includegraphics[width=4.54cm]{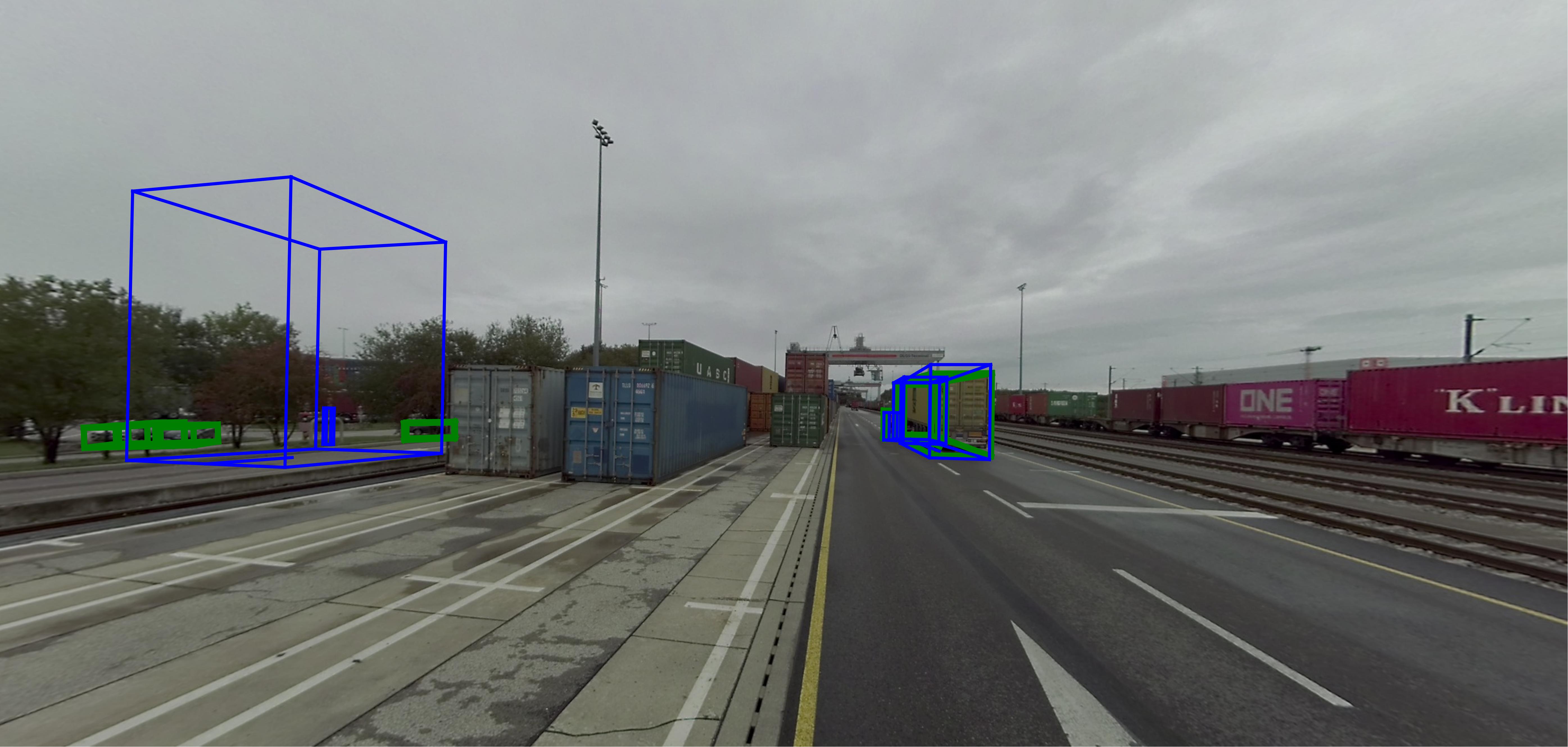}
    \end{subfigure}
    \\ \vspace{8pt}
    \begin{subfigure}{0.325\linewidth}
      \centering
      \includegraphics[height=4.54cm,angle=-90,origin=c]{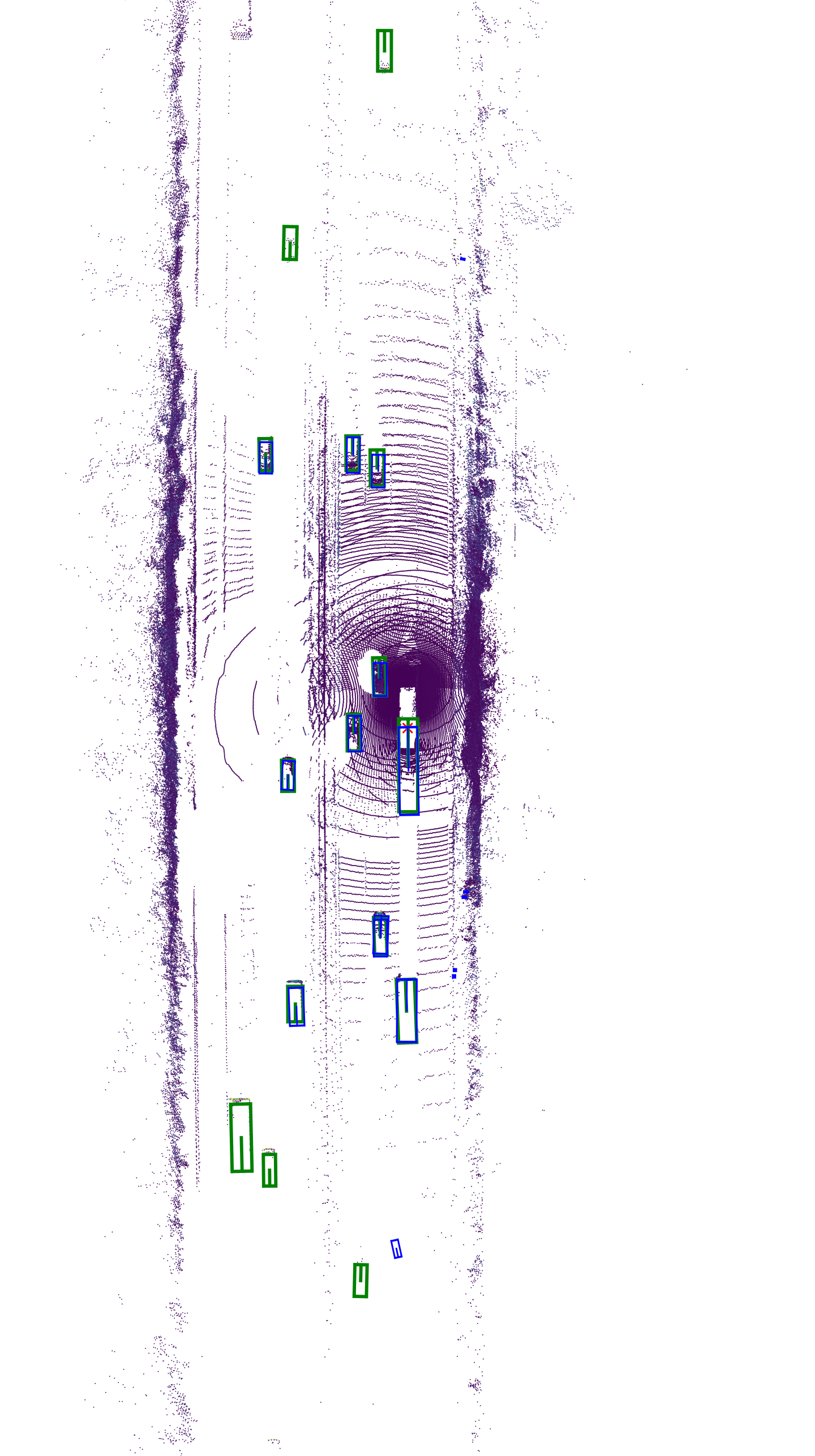}\vspace{-24pt}
      \caption{night}
    \end{subfigure}
    \hfill
    \begin{subfigure}{0.325\linewidth}
      \centering
      \includegraphics[height=4.54cm,angle=-90,origin=c]{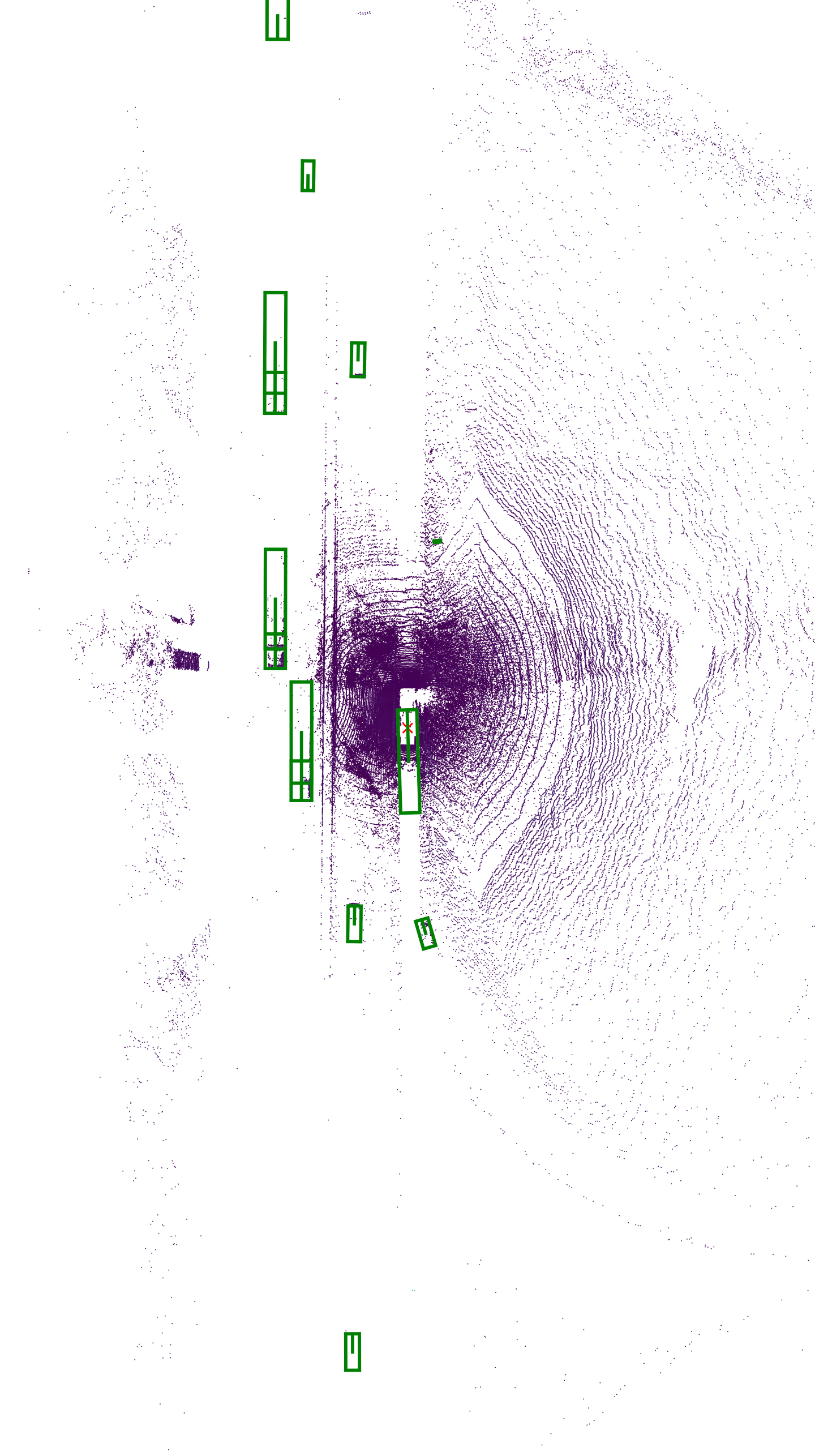}\vspace{-24pt}
      \caption{snow}
    \end{subfigure}
    \hfill
    \begin{subfigure}{0.325\linewidth}
      \centering
      \includegraphics[height=4.54cm,angle=-90,origin=c]{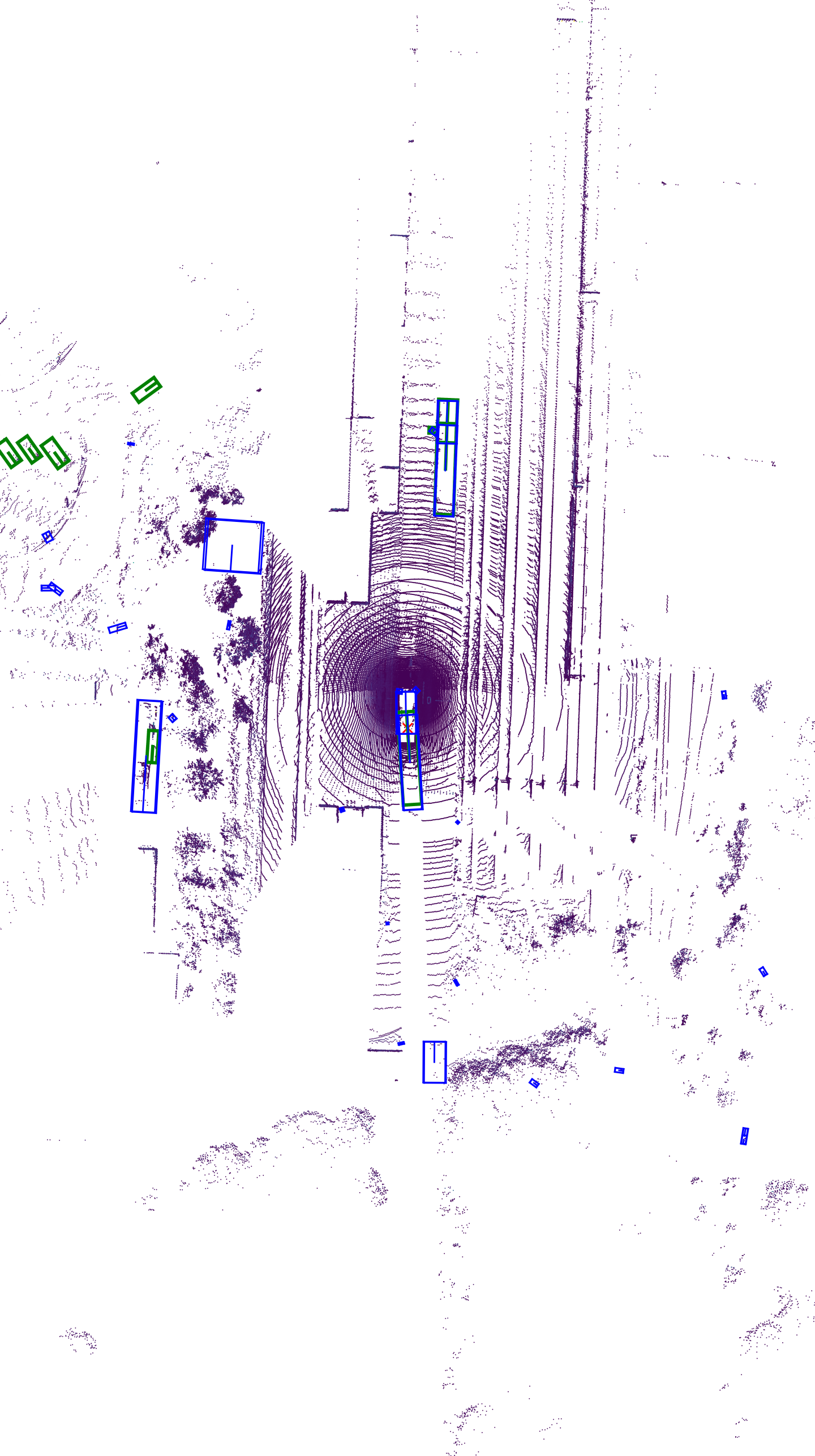}\vspace{-24pt}
      \caption{terminal}
    \end{subfigure}

    \caption{Example visualizations of the CenterPoint model on the validation split of the \MANDataset{} dataset v1.0. The front left camera is shown on the top and the fused lidar point cloud on the bottom. Model predictions are shown in \textcolor{blue}{blue} and the ground truth in \textcolor{green}{green}.}
    \label{fig:lidar_examples}
\end{figure}

In the following, detailed results of the different models are provided for every sensor modality. Table~\ref{tab:camera_class_results} - \ref{tab:lidar_class_results} show the class-specific results per sensor modality, while Table~\ref{tab:range_results} lists the object detection results over different range bins. Table~\ref{tab:tag_results} shows the detection results by scene tags and Figure~\ref{fig:camera_examples}  - \ref{fig:lidar_examples} provides qualitative results under various environmental conditions.

Observations show that the camera model struggles to perform well on minority classes and performs generally better on larger than smaller objects. The results show lots of false positives (FP) and the accumulation of bounding boxes on drivable areas. Besides that, the camera model's performance decreases with increasing distance and performs significantly worse in the dark or tunnel environments. 

The radar model, on the other hand, struggles to detect non-metallic objects but shows promising results on the majority classes. The results of the radar experiments also suggest that the accurate prediction of object orientations is most challenging and the velocity estimation seems to be off due to the utilization of the relative radial velocity instead of the ego-motion compensated velocity measurements. Furthermore, the radar-based detection performs significantly worse in the terminal environment and environments with a lot of radar reflections. 

Lastly, the results of the CenterPoint model training show that the detection quality of diverse object classes (like 'animal', 'other vehicle', or 'bus') is insufficient, while more homogeneous object classes (like 'traffic sign', 'traffic cone', or 'pedestrian') have a higher detection score. Moreover, the size of the objects or the number of objects in the dataset are no sufficient predictors for the detection quality, as shown by the 'truck' class. In addition, the results suggest that the detection quality decreases with increasing distance and that rainy or foggy weather conditions negatively affect the model's performance. The model also performs significantly worse in tunnels or rural areas. However, safe autonomous trucking requires robust perception under all environmental conditions, which is why we release \MANDataset{} to promote research in this area.

\begin{table}[h]
    \caption{Results of the camera 3D object detection task per detection class on the test split of the \MANDataset{} dataset v1.0.}
    \label{tab:camera_class_results}
    \centering
    \begin{tabular}{lcccccc}
        \toprule
                       & \multicolumn{6}{c}{PETR}         \\
                       & AP   & ATE  & ASE  & AOE  & AVE  & AAE  \\
                      
        \midrule
        car            & 0.03 & 1.26 & 0.22 & 0.20 & 4.18 & 0.12 \\
        truck          & 0.01 & 1.27 & 0.35 & 0.11 & 2.17 & 0.20 \\
        bus            & 0.00 & 1.00 & 1.00 & 1.00 & 1.00 & 1.00 \\
        trailer        & 0.19 & 0.67 & 0.19 & 0.04 & 1.84 & 0.12 \\
        other\_vehicle & 0.00 & 1.00 & 1.00 & 1.00 & 1.00 & 1.00 \\
        pedestrian     & 0.00 & 1.25 & 0.48 & 1.19 & 0.30 & 0.63 \\
        motorcycle     & 0.00 & 1.00 & 1.00 & 1.00 & 1.00 & 1.00 \\
        bicycle        & 0.00 & 1.00 & 1.00 & 1.00 & 1.00 & 1.00 \\
        traffic\_cone  & 0.03 & 1.31 & 0.51 &    - &    - &    - \\
        barrier        & 0.01 & 1.46 & 0.90 & 0.07 &    - &    - \\
        animal         & 0.00 & 1.00 & 1.00 & 1.00 & 1.00 &    - \\
        traffic\_sign  & 0.00 & 1.29 & 0.61 & 0.50 &    - &  0.02 \\
        \bottomrule
    \end{tabular}
\end{table}

\begin{table}[h]
    \caption{Results of the radar 3D object detection task per detection class on the test split of the \MANDataset{} dataset v1.0.}
    \label{tab:radar_class_results}
    \centering
    \begin{tabular}{lcccccc}
        \toprule
                       & \multicolumn{6}{c}{RadarGNN}         \\
                       & AP   & ATE  & ASE  & AOE  & AVE  & AAE  \\
                      
        \midrule
		car            & 0.36 & 0.53 & 0.22 & 1.48 & 26.50 & 0.02 \\
        truck          & 0.20 & 0.87 & 0.42 & 1.61 & 20.67 & 0.02 \\
        bus            & 0.00 & 1.00 & 1.00 & 1.00 & 1.00  & 1.00 \\
        trailer        & 0.27 & 0.66 & 0.15 & 1.09 & 18.86 & 0.02 \\
        other\_vehicle & 0.00 & 1.00 & 1.00 & 1.00 & 1.00  & 1.00 \\
        pedestrian     & 0.00 & 1.00 & 1.00 & 1.00 & 1.00  & 1.00 \\
        motorcycle     & 0.00 & 1.00 & 1.00 & 1.00 & 1.00  & 1.00 \\
        bicycle        & 0.00 & 1.00 & 1.00 & 1.00 & 1.00  & 1.00 \\
        traffic\_cone  & 0.00 & 1.00 & 1.00 &    - &    -  &    - \\
        barrier        & 0.00 & 1.00 & 1.00 & 1.00 &    -  &    - \\
        animal         & 0.00 & 1.00 & 1.00 & 1.00 & 1.00  &    - \\
        traffic\_sign  & 0.02 & 0.64 & 0.92 & 1.27 &    -  & 0.08 \\
        \bottomrule
    \end{tabular}
\end{table}

\begin{table}[h]
    \caption{Results of the lidar 3D object detection task per detection class on the test split of the \MANDataset{} dataset v1.0.}
    \label{tab:lidar_class_results}
    \centering
    \begin{tabular}{lcccccc}
        \toprule
                       & \multicolumn{6}{c}{CenterPoint}         \\
                       & AP   & ATE  & ASE  & AOE  & AVE  & AAE  \\
                      
        \midrule
        car            & 0.33 & 0.29 & 0.15 & 0.07 & 3.95 & 0.17 \\
        truck          & 0.23 & 0.42 & 0.16 & 0.03 & 2.62 & 0.39 \\
        bus            & 0.06 & 0.49 & 0.11 & 0.18 & 4.28 & 0.01 \\
        trailer        & 0.34 & 0.42 & 0.15 & 0.03 & 3.74 & 0.37 \\
        other\_vehicle & 0.14 & 0.65 & 0.34 & 0.15 & 0.85 & 0.32 \\
        pedestrian     & 0.45 & 0.29 & 0.39 & 0.81 & 0.43 & 0.37 \\
        motorcycle     & 0.22 & 0.33 & 0.22 & 0.08 & 5.57 & 0.03 \\
        bicycle        & 0.06 & 0.27 & 0.27 & 0.46 & 2.13 & 0.13 \\
        traffic\_cone  & 0.55 & 0.28 & 0.37 &    - &    - &    - \\
        barrier        & 0.42 & 0.19 & 0.62 & 0.03 &    - &    - \\
        animal         & 0.00 & 1.00 & 1.00 & 1.00 & 1.00 &    - \\
        traffic\_sign  & 0.42 & 0.28 & 0.44 & 0.21 &    - & 0.03 \\
        \bottomrule
    \end{tabular}
\end{table}

\begin{table}[h]
    \caption{3D object detection results (mAP / NDS) by range on the \MANDataset{} test set v1.0.}
    \label{tab:range_results}
    \centering
    \begin{tabular}{llcccc}
        \toprule
                    & Modality & \SI{0}{\metre} - \SI{25}{\metre} & \SI{0}{\metre} - \SI{50}{\metre} & \SI{0}{\metre} - \SI{100}{\metre} & \SI{0}{\metre} - \SI{150}{\metre} \\
        \midrule
        PETR & camera & 0.06 / 0.13       & 0.04 / 0.13        & 0.03 / 0.12        & 0.02 / 0.12        \\
        RadarGNN & radar & 0.09 / 0.12       & 0.08 / 0.12        & 0.08 / 0.11        & 0.07 / 0.11        \\
        CenterPoint & lidar & 0.48 / 0.52       & 0.40 / 0.48        & 0.30 / 0.43        & 0.27 / 0.41        \\
        \bottomrule
    \end{tabular}
\end{table}

\begin{table}[t]
    \caption{Results (mAP / NDS) of the 3D object detection task by scene tag on the test split of the \MANDataset{} dataset v1.0.}
    \label{tab:tag_results}
    \centering
    \begin{tabular}{clccc}
        \toprule
                                      &                 & PETR & RadarGNN & CenterPoint \\
                                      & modality        & camera & radar & lidar \\
        \midrule
        \multirow{7}{*}{\STAB{\rotatebox[origin=c]{90}{weather}}}      & clear           & 0.02 / 0.11 & 0.05 / 0.10 & 0.30 / 0.42            \\
                                      & overcast        & 0.02 / 0.11 & 0.09 / 0.12 & 0.19 / 0.36            \\
                                      & fog             & 0.05 / 0.13 & 0.15 / 0.15 & 0.15 / 0.20            \\
                                      & other\_weather  & - / - & - / - & - / -       \\
                                      & rain            & 0.03 / 0.11 & 0.09 / 0.12 & 0.16 / 0.23            \\
                                      & snow            & - / - & - / - & - / -       \\
                                      & hail            & - / - & - / - & - / -       \\
        \midrule
        \multirow{7}{*}{\STAB{\rotatebox[origin=c]{90}{area}}}         & terminal        & 0.02 / 0.09 & 0.00 / 0.04 & 0.30 / 0.32            \\
                                      & parking         & - / - & - / - & - / -       \\
                                      & highway         & 0.02 / 0.11 & 0.09 / 0.12 & 0.19 / 0.31            \\
                                      & city            & 0.02 / 0.07 & 0.02 / 0.06 & 0.18 / 0.29            \\
                                      & residential     & 0.01 / 0.06 & 0.03 / 0.07 & 0.12 / 0.28            \\
                                      & rural           & 0.03 / 0.10 & 0.07 / 0.11 & 0.06 / 0.15            \\
                                      & other\_area     & - / - & - / - & - / -       \\
        \midrule
        \multirow{4}{*}{\STAB{\rotatebox[origin=c]{90}{daytime}}}      & morning         & 0.02 / 0.11 & 0.08 / 0.11 & 0.17 / 0.36            \\
                                      & noon            & 0.02 / 0.12 & 0.06 / 0.11 & 0.33 / 0.44            \\
                                      & evening         & - / - & - / - & - / -       \\
                                      & night           & 0.02 / 0.10 & 0.06 / 0.09 & 0.17 / 0.24            \\
        \midrule
        \multirow{3}{*}{\STAB{\rotatebox[origin=c]{90}{season}}}       & spring          & - / - & - / - & - / -          \\
                                      & summer          & 0.02 / 0.12 & 0.07 / 0.11 & 0.34 / 0.42            \\
                                      & autumn          & 0.02 / 0.11 & 0.08 / 0.11 & 0.20 / 0.35            \\
                                      & winter          & 0.03 / 0.10 & 0.05 / 0.10 & 0.12 / 0.26            \\
        \midrule
        \multirow{5}{*}{\STAB{\rotatebox[origin=c]{90}{lighting}}}     & illuminated     & 0.02 / 0.12 & 0.07 / 0.11 & 0.26 / 0.41            \\
                                      & twilight        & 0.05 / 0.09 & 0.06 / 0.10 & 0.19 / 0.27            \\
                                      & dark            & 0.01 / 0.09 & 0.02 / 0.09 & 0.18 / 0.24            \\
                                      & glare           & 0.02 / 0.11 & 0.09 / 0.12 & 0.16 / 0.23            \\
                                      & other\_lighting & - / - & - / - & - / -       \\
        \midrule
        \multirow{5}{*}{\STAB{\rotatebox[origin=c]{90}{structure}}}    & regular         & 0.02 / 0.12 & 0.06 / 0.10 & 0.29 / 0.39            \\
                                      & underpass       & 0.04 / 0.12 & 0.11 / 0.13 & 0.13 / 0.28            \\
                                      & bridge          & 0.03 / 0.11 & 0.09 / 0.12 & 0.14 / 0.30            \\
                                      & overpass        & 0.03 / 0.12 & 0.09 / 0.12 & 0.24 / 0.31            \\
                                      & tunnel          & 0.03 / 0.08 & 0.07 / 0.09 & 0.12 / 0.16            \\
        \midrule
        \multirow{2}{*}{\STAB{\rotatebox[origin=c]{90}{const}}} & unchanged       & 0.02 / 0.12 & 0.07 / 0.11 & 0.25 / 0.40            \\
                                      & roadworks       & 0.04 / 0.12 & 0.10 / 0.11 & 0.23 / 0.28            \\
        \bottomrule
    \end{tabular}
\end{table}

\newpage

% End

\end{document}